\documentclass[shortpaper]{clv3_no_hints}
\usepackage{hyperref}
\usepackage{xcolor}
\usepackage{graphicx}
\usepackage{subcaption}
\usepackage{soul}
\usepackage[disable]{todonotes}
\usepackage{tabularx}
\definecolor{darkblue}{rgb}{0, 0, 0.5}
\hypersetup{colorlinks=true,citecolor=darkblue, linkcolor=darkblue, urlcolor=darkblue}

\newcommand\bert{\textsc{BERT}}
\newcommand\mlmsquad{\textsc{Mlm-SQuAD}}
\newcommand\mlmmarco{\textsc{Mlm-MSMarco}}
\newcommand\qasquad{\textsc{Qa-SQuAD-1}}
\newcommand\qasquadbig{\textsc{Qa-SQuAD-2}}

\newcommand\rankmarco{\textsc{Rank-MSMarco}}
\newcommand\ner{\textsc{Ner-CoNLL}}
\newcommand\squadprobe{\texttt{Squad}}
\newcommand\trexprobe{\texttt{T-REx}}
\newcommand\concpetprobe{\texttt{ConceptNet}}
\newcommand\greprobe{\texttt{Google-RE}}
\newcommand{\mpara}[1]{\medskip\noindent{\bf #1}}

\runningtitle{BERTnesia: Investigating the capture and forgetting of knowledge in BERT}

\runningauthor{Jonas Wallat}

\begin{document}

\title{BERTnesia: Investigating the capture and forgetting of knowledge in BERT}

\author{Jonas Wallat}
\affil{L3S Research Center\\Hannover, Germany\\\texttt{wallat@l3s.de}}

\author{Jaspreet Singh}
\affil{L3S Research Center\\Hannover, Germany\\\texttt{singh@l3s.de}}

\author{Avishek Anand}
\affil{L3S Research Center\\Hannover, Germany\\\texttt{anand@l3s.de}}

\maketitle

\begin{abstract}
Probing complex language models has recently revealed several insights into linguistic and semantic patterns found in the learned representations. In this article, we probe \bert{} specifically to understand and measure the relational knowledge it captures in its parametric memory. While probing for linguistic understanding is commonly applied to all layers of \bert{} as well as fine-tuned models, this has not been done for factual knowledge. We utilize existing knowledge base completion tasks (LAMA) to probe every layer of pre-trained as well as fine-tuned \bert{} models (ranking, question answering, NER). 
Our findings show that knowledge is not just contained in \bert{}’s final layers. Intermediate layers contribute a significant amount (17-60\%) to the total knowledge found. Probing intermediate layers also reveals how different types of knowledge emerge at varying rates.
When \bert{} is fine-tuned, relational knowledge is forgotten. The extent of forgetting is impacted by the fine-tuning objective and the training data. We found that ranking models forget the least and retain more knowledge in their final layer compared to masked language modeling and question-answering. However, masked language modeling performed the best at acquiring new knowledge from the training data. When it comes to learning facts, we found that capacity and fact density are key factors. We hope this initial work will spur further research into understanding the parametric memory of language models and the effect of training objectives on factual knowledge. The code to repeat the experiments is publicly available on GitHub\footnote{https://github.com/jwallat/knowledge-probing}. \todo{We need to mention the first paper and state how we extended the work somewhere}
\end{abstract}

\section{Introduction}

Large pre-trained language models like \bert{}~\cite{devlin2018bert} have heralded an \textit{ImageNet} moment for NLP\footnote{https://thegradient.pub/nlp-imagenet/} with not only significant improvements made to traditional tasks such as question answering and machine translation but also in the new areas such as knowledge base completion.
The \bert{} family of language models essentially showcase the need for over-parameterization and modelling long-term interaction in textual input for improved language understanding.
However, improved performance for such models comes at the expense of reduced interpretability.

Language models have been shown to understand linguistic information like local syntax, long-range semantics or even compositional reasoning.
A recent approach called \emph{probing} is one approach to inspect the inner workings of \bert{} and other complex language models for better interpretability~\cite{dasgupta2018evaluating,ettinger-etal-2018-assessing, tenney2019you}. 
In general, probing is a procedure that tests if a specific linguistic information can be decoded from a model's latent embeddings.
Corresponding to the desired linguistic information under examination, probing inputs are constructed to explicitly test the presence of linguistic patterns.

\begin{figure}
    \centering
    \includegraphics[width=0.90\textwidth]{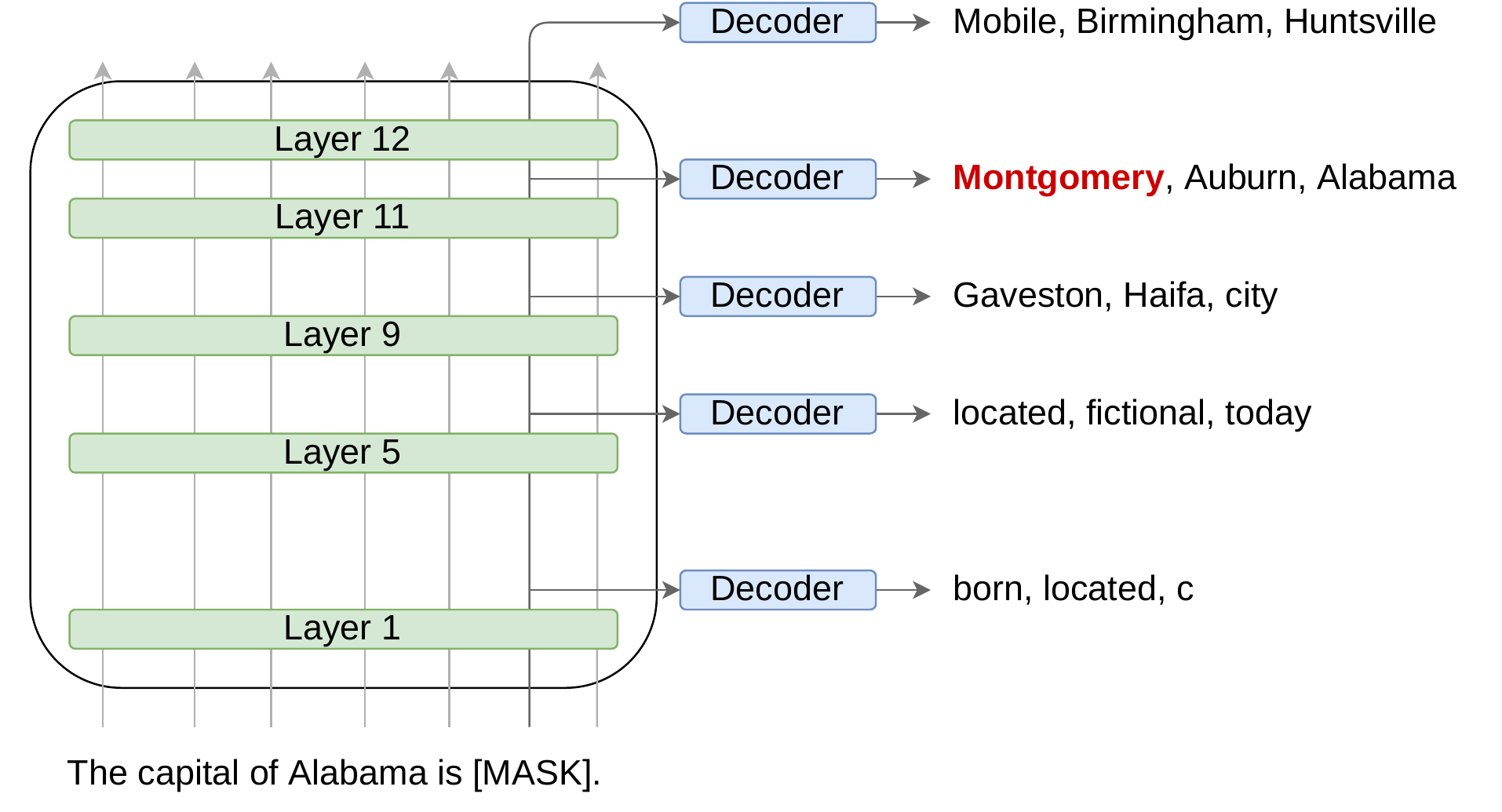}
    \caption{Overview of \bert{} with our proposed layer-wise probing procedure. Given a masked word, the masked language modeling head (i.e., the decoder) will use the \bert{} embeddings to produce a probability distribution over the vocabulary. The figure shows the top 3 tokens for a selection of layers with the correct answer in boldface. We find not all knowledge to be accessible in the last layers -- a significant amount of facts as well as some relationships are better captured at intermediary layers.}
    \label{fig:money_fig_probing_procedure}
\end{figure}

More interestingly, recent studies have confirmed that \bert{} also acquires factual and relational knowledge from their training process.
Seminal work by~\citet{petroni2019language} probed \bert{} and other language models for relational knowledge (e.g., \textit{Trump} \textsf{is the president of} the \textit{USA}) in order to determine the potential of using language models as automatic knowledge bases. 
Their approach converted queries in the knowledge base (KB) completion task of predicting arguments or relations from a KB triple into a natural language cloze task, for example, \texttt{[MASK]} \textsf{is the president of} the \textit{USA}.
This is done to make the query compatible with the pre-training masked language modeling (MLM) objective. 
By considering multiple probe sets (also called as the LAMA probes), they consequently showed that a reasonable amount of knowledge is captured in \bert{}. 
As a consequence, factual knowledge stored in the parametric memory of \bert{} models can be used for knowledge-intensive tasks like question answering and fact checking without the need of additional context~\citep{roberts-etal-2020-much, Lewis2021PAQ6M}.

However, many questions about the information content in the parametric memory of \bert{} are unanswered. 
Firstly, the existing probing methodology focuses on the final layer of \bert{} and could underestimate the knowledge contained in the lower layers. This prompts the question -- \textit{Is there more knowledge in \bert{} than what is reported? }
Secondly, the effect of fine-tuning on the relational knowledge is not clearly understood. In other words -- \emph{What happens to relational knowledge when \bert{} is fine-tuned for other tasks?}
Finally, we study the knowledge evolution through the layers of the parametric memory of \bert{}, that is, we attempt to understand \emph{how knowledge is gained and/or forgotten through the layers of \bert{}?}
To the best of our knowledge, the effect and efficiency of training tasks other than language modeling on factual knowledge is largely unexplored. This is the extended version of our initial work~\citep{singh-etal-2020-bertnesia} that appeared at the BlackboxNLP 2020 workshop\footnote{https://blackboxnlp.github.io/2020/}. Whereas the prior work was able to show \textit{how} knowledge evolves and is forgotten, this extended version aims to detail \textit{why} these effects occur. 

\subsection{Our contribution} 
To improve our understanding of the parametric memory, we extensively study the emergence of knowledge through the layers in \bert{} by devising a procedure to estimate knowledge contained in every layer and not just the last (as done by~\citet{petroni2019language}).
Our probing procedure is roughly sketched in Figure~\ref{fig:money_fig_probing_procedure} that shows the use of a light-weight decoder for the representations at each \bert{} layer.

\mpara{Knowledge containment.} We show that existing studies that observing only the final layer underestimate the amount of factual knowledge in \bert{}'s parametric memory.
Specifically, we find that a substantial amount of knowledge ($\sim24\%$) is stored in the intermediate layers (Section~\ref{sec:intermediate}).

\mpara{Knowledge evolution.} Additionally, we also provide insights into how relational knowledge emerges through \bert{}'s layers.
We find that not all relational knowledge is captured gradually through the layers with 15\% of relationship types essentially doubling in the last layer and 7\% of relationship types being maximally captured in an intermediate layer (Section~\ref{sec:evolution}). This is further evidence that not all knowledge is accessible in the last layer. 

\mpara{Fine-tuning and forgetting.} We find that fine-tuning always causes forgetting (Section~\ref{sec:finetuning}). 
When the size of the dataset is fixed and training objective varies, the ranking model (\rankmarco{} in our experiments) forgets less than the QA model (35\% vs. 53\%). 
When it comes to acquiring new knowledge from fine-tuning, we find MLM to be more effective than QA and ranking (26.5\% vs. 17.25\% and 16.5\%).

\mpara{Impact of Training data.} We find that the dataset size does not play a major role when the training objective is fixed as MLM. Fine-tuning on a larger dataset does not lead to less forgetting (Section~\ref{sec:data_size}).
As the capacity to store factual knowledge is limited, training on larger amounts of new data will result in more previous knowledge being forgotten. We also observe the density of factual information in the training data to be a factor when it comes to retaining knowledge.

\subsection{Insights and implications.}
We believe that understanding how and under what conditions knowledge is acquired, stored, and forgotten in the parametric memory of \bert{} would help develop better knowledge-intensive models.
On one hand, our findings can help infuse more factual knowledge into the parametric memory is by enriching the pre-training data with additional knowledge retrieved from a large textual corpus~\citep{Guu2020REALMRL}.
On the other hand, our findings can also help design novel retrieval strategies for extracting relevant factual knowledge from the \bert{}'s parametric memory.

\section{Related Work}

In this section, we survey previous work on probing language models (LMs) as well as acquiring and forgetting (factual) knowledge. For the related work on probing, we particularly focus on contextual embeddings learned by \bert{}. Probes have been designed for both static and contextualized word representations. Static embeddings refer to non-contextual embeddings such as GloVe~\cite{pennington2014glove}. 
For the static case, the reader can refer to this survey by ~\citet{Belinkov_2019}. 
In the following, we detail probing tasks for contextualized embeddings from language models.



\subsection{Probing for syntax, semantics, and grammar}


Initial work on probing dealt with linguistic pattern detection. 
\citet{Peters2018DissectingCW} investigated the ability of various neural network architectures that learn contextualized word representations to capture local syntax and long-range semantics like co-reference resolution while
\citet{dasgupta2018evaluating,ettinger-etal-2018-assessing} probed language models for compositional reasoning. 


\citet{McCoy_2019, goldberg2019assessing} found that \bert{} is able to effectively learn syntactic heuristics with natural language inference specific probes.  \citet{tenney2019bert, liu2019linguistic, jawahar-etal-2019-bert} investigated \bert{} layer-by-layer for various syntactic and semantic patterns like part-of-speech, named entity recognition, co-reference resolution, entity type prediction, semantic role labeling, etc.
They all found that basic linguistic patterns like part of speech emerge at the lower layers.
However, there is no consensus with regards to semantics with somewhat conflicting findings (equally spread vs. final layer~\cite{jawahar-etal-2019-bert}). 
~\citet{kovaleva-etal-2019-revealing} found that the last layers of fine-tuned \bert{} contain the most amount of task-specific knowledge. \citet{vanaken_2019} showed the same result for fined-tuned QA \bert{} with specially designed probes. They found that the lower and intermediary layers of the QA model were better suited to linguistic subtasks associated with QA.
For a more comprehensive survey on probing for linguistic information we point the reader to~\cite{rogers2020primer}.



Our work is similar to these studies in terms of setup. 
In particular, our probes function on the sentence level and are applied to each layer of a pre-trained \bert{} model as well as \bert{} fine-tuned on several tasks. 
However, we do not focus on detecting linguistic patterns and focus on relational and factual knowledge - how knowledge is acquired, forgotten and the effects of training tasks.



\subsection{Probing for knowledge}
\label{sec:know_probe}

In parallel, there have been investigations into probing for factual and world knowledge. 
Initially, \citet{petroni2019language} found that LMs like \bert{} can be directly used for the task of knowledge base completion since they are able to memorize more facts than some automatic knowledge bases. 
They created cloze statement tasks for factual and commonsense knowledge (LAMA) and measured cloze-task performance as a proxy for the knowledge contained. 
However, using the same probing framework, \citet{poerner-etal-2020-e} showed that this factoid knowledge is influenced by surface-level stereotypes of words. For example, \bert{} tends to predict a typically French sounding name to be a French citizen. Originally, \citet{petroni2019language} manually formulated cloze statements and quite a few recent works have been on designing templates (or prompts) that are more effective at eliciting knowledge from \bert{}. In that line of work, \citet{bouraoui2019inducing} mined Wikipedia for sentences mentioning a fact and evaluated them by using \bert{} to predict the masked object. With LPAQA, \citet{Jiang2019HowCW} propose a different set of templates, that improved \bert{}'s performance on the LAMA data. Besides mining based approaches, \citet{Jiang2019HowCW} also used paraphrasing methods to introduce more diversity and constructed the final set of templates by ensembling. Another recent method is AutoPrompt by \citet{shin-etal-2020-autoprompt}, who devised a gradient-based search strategy for fixed-size templates. The templates generated with AutoPrompt are often not human understandable (Original: The native language of Bjork is [MASK] -> Bjorkneau optionally fluent!? traditional [MASK].), however allow for eliciting even more knowledge from \bert{}. Most recently, \citet{Zhong2021FactualPI} proposed OptiPrompt, which further improved performance on the LAMA data. While other approaches optimized the discrete input templates (i.e., words), OptiPrompt optimized the continuous embedding space. \citet{Zhong2021FactualPI} found this to be highly effective, yet even less interpretable to humans. Related to the works on designing better templates, \citet{Elazar2021MeasuringAI} study how much \bert{}'s predictions change given different paraphrases of the same fact (PARAREL), finding the outputs to lack consistency. Similarly, to overcome inconsistencies, \citet{Kassner2021EnrichingAM} proposed adding both a persistent memory component, keeping track of their model's beliefs and a SAT solver, that checks for clashing believes. Extending QA model with these components resulted in improved accuracy and consistency. 

The majority of the previous research on factual knowledge in language models was limited to the English language. \citet{kassner-etal-2021-multilingual} probed the multilingual \bert{} (m\bert{}) for its factual knowledge. While m\bert{} performed close to the monolingual (English) \bert{} on languages as English, Spanish and French, it performed significantly worse on others such as Japanese or Thai. This discrepancy between individual languages suggests that facts and entity knowledge is not stored independent from the languages.

Tangentially to the work on the LAMA data, \citet{forbes2019neural} investigated \bert{}'s awareness of the world. They devised object property and action probes to estimate \bert{}'s ability to reason about the physical world. They found that \bert{} is relatively incapable of such reasoning but is able to memorize some properties of real-world objects. This investigation tested commonsense spatial reasoning rather than pure factoid knowledge.

Following the finding of \citet{Geva2020TransformerFL}, that feed forward layers are key-value memories, \citet{Dai2021KnowledgeNI} investigated where factual knowledge is located in transformer blocks. Using integrated gradients \citep{Sundararajan2017AxiomaticAF}, they proposed a knowledge attribution method that selects a small number of neurons ($\sim$6 on average) that significantly hurt the fact performance when activations are suppressed.

In this work, we utilize the LAMA probing data, but rather than proposing new templates, we aim to develop a better understanding of how the knowledge is stored and evolves in \bert{} and its fine-tuned variants. Here, we are more interested in relative differences, in which layers resides the most amount of knowledge and how fine-tuning effects factual knowledge in language models. To this end, we adapt the layer-wise probing methodology often employed for linguistic pattern detection by ~\citet{vanaken_2019,tenney2019bert,liu2019linguistic} for the probe tasks suggested in~\citet{petroni2019language}.

\subsection{Learning and forgetting}
The question of how we can inject more knowledge into the parametric memory of language models has been studied in many works. One prominent way of learning more factual knowledge is masking entities instead of random words in pre-training. This was done by \citet{sun2019ERNIEER} for their \textsc{ERNIE}-model, which improved for example named entity recognition and knowledge inference. \citet{Guu2020REALMRL} proposed adding a latent knowledge retriever to the pre-training process, which will extend the context with additional knowledge derived from a textual corpus. The latter pre-training procedure is also commonly used to improve the performance of closed-book question-answering (CBQA) models \citep{roberts-etal-2020-much, Lewis2021PAQ6M}. CBQA is highly related to the probing considered in this article: both settings require the model to produce the correct answer directly from their parametric memory, without access to outside sources. More generally, \citet{kassner-etal-2020-symbolic-reasoners} used a synthetic dataset to investigate what effects make \bert{} remember facts, finding that schema conformity and frequency are important factors. 

Besides learning facts from training, such knowledge can also be directly infused into the embeddings. \citet{poerner-etal-2020-e} align Wikipedia2Vec entity vectors \citep{yamada-etal-2016-joint} with \bert{} and call their new model entity-enhanced \bert{} (\textsc{E-BERT}). This entity-enhanced version outperforms the regular \bert{} on cloze-style question answering, relation classification and entity linking. Besides word vectors, also knowledge graphs have been used to inject knowledge into language models with a variety of techniques \citep{Liu2020KBERTEL, Wang2020KAdapterIK, Peters2019KnowledgeEC, he-etal-2020-bert}. 

While learning and infusing knowledge is well studied for the task of language modeling, it is largely unexplored for other down-stream tasks (e.g., question-answering, ranking). However, one thing that has been studied is the effect that models lose the ability to do their pre-training task when being fine-tuned. This sequential learning problem has also been referred to as catastrophic inference \citep{mccloskey1989catastrophic} or catastrophic forgetting \citep{kirkpatrick2017overcoming, mosbach2020stability}. This is an ongoing problem as general AI models will need to be able to learn and perform multiple tasks without forgetting how to perform the former. \citet{kirkpatrick2017overcoming} suggest slowing down learning of important weights for the initial task, whereas \citet{mosbach2020stability} investigated the role of catastrophic forgetting on fine-tuning stability, finding that the last few layers are usually replaced with task-specific knowledge. If and how factual knowledge is affected by catastrophic forgetting is largely unexplored.
In this work, we will take a first step into understanding how fine-tuning tasks and data effect both the learning and forgetting of factual knowledge.

There has been recent work on forgetting in attention modules when dealing with long contexts \citep{Child2019GeneratingLS, Schlag2021LinearTA}. When the capacity is exhausted, previous information is dropped in favor of the more recent context. By learning which parts of the context are important and systematically expiring ones faster that are not, \citet{Sukhbaatar2021NotAM} reduced the necessary capacity to attend to extremely long sequences. The work in this article is different in that we use probes without contexts as we want to retrieve knowledge from the model's parametric memory. Therefore, we do not investigate the attention modules but the learned embeddings.    


\section{Experimental Setup}
\label{sec:setup}

\subsection{Models}
\label{sec:models}

\bert{} is a bidirectional text encoder built by stacking several transformer layers. 
\bert{} is often pre-trained with two tasks: next sentence classification and masked language modeling (MLM). 
MLM is cast as a classification task over all tokens in the vocabulary.
It is realized by training a decoder that takes as input the mask token embedding and outputs a probability distribution over vocabulary tokens. In our experiments we used \bert{} base (12 layers) pre-trained on the BooksCorpus ~\cite{zhu2015bookCorpus} and English Wikipedia. 
We use this model for fine-tuning to keep comparisons consistent. 
Henceforth, we refer to pre-trained \bert{} as just \bert{}. Table~\ref{tab:model_selection} details all models used in our experiments.

    
    
    
    
    

\begin{table}[h]
\centering
\begin{tabular}{lll}
\hline
\textbf{Name} & \textbf{Task}                                       & \textbf{Dataset}               \\ \hline
BERT          & MLM + NSP & BooksCorpus, Wikipedia \\
NER-CoNLL     & named entity recognition                            & CoNLL-2003                     \\
QA-SQuAD-1    & QA (span prediction)                                & SQuAD 1.1                      \\
QA-SQuAD-2    & QA (span prediction + unanswerable)                                & SQuAD 2                        \\
RANK-MSMarco  & passage re-ranking                                  & MSMarco                        \\
MLM-MSMarco   & masked language modeling                            & MSMarco                        \\
MLM-SQuAD     & masked language modeling                            & SQuAD 1.1                      \\ \hline
\end{tabular}
\caption{List of the models used in our experiments. Additional information on training parameters and the datasets can be found in the Appendix.}
\label{tab:model_selection}
\end{table}

When fine-tuning, our goal was to not only achieve good performance but also to minimize the number of extra parameters added. More parameters outside \bert{} may increase the chance of knowledge being stored elsewhere leading to unreliable measurement. We used the Huggingface transformers library \cite{Wolf2019HuggingFacesTS} for implementing all models in our experiments. More details on hyperparameters and training can be found in the Appendix.

\subsection{Knowledge probes}
\label{sec:probes}
We utilized the existing suite of LAMA knowledge probes suggested in~\cite{petroni2019language}\footnote{ https://github.com/facebookresearch/LAMA} for our experiments. Table~\ref{tab:kp_details} briefly summarizes the key details. The probes are designed as cloze statements and limited to single token factual knowledge. Multi-word entities and relations are not included.

\begin{table*}[h]
\centering
\small
\begin{tabular}{lccll}
\hline
\textbf{Probe set} & \textbf{\#Rels} & \textbf{\#Instances} & \textbf{Example} & \textbf{Answer} \\
\hline
\concpetprobe{} & - & 12514 & Rocks are [MASK]. & {solid} \\
\trexprobe{} & 41 & \ 34017 & The capital of Germany is [MASK]. & {Berlin} \\
\greprobe{} & 3 & 5528 & Eyolf Kleven was born in [MASK]. & {Copenhagen} \\
\squadprobe{} & - & 305 & Nathan Alterman was a [MASK]. & {Poet} \\
\hline
\end{tabular}
\caption{Knowledge probes used in the experiments. \citet{petroni2019language} subsampled \concpetprobe{} \cite{speer2012representing}, \trexprobe{} \cite{elsahar2019t}, \greprobe{} \cite{orr201350} and \squadprobe{} \cite{rajpurkar2016squad}.}
\label{tab:kp_details}
\end{table*}

Each probe in LAMA is constructed to test a specific relation or type of relational knowledge.  \concpetprobe{} is designed to test for general conceptual knowledge since it masks single token objects from randomly sampled sentences whereas \trexprobe{} consists of hundreds of sentences for 41 specific relationship types like \textit{member of} and  \textit{language spoken}. 
\greprobe{} tests for 3 specific types of factual knowledge related to people: place-of-birth (2937), date-of-birth (1825), and place-of-death (766 instances). The date-of-birth is a strict numeric prediction that is not covered by \trexprobe{}. 
Finally, \squadprobe{} uses context insensitive questions from SQuAD that have been manually rewritten to cloze-style statements. 
Note that this is the same dataset used to train \qasquad{} and \qasquadbig{}. 


\subsection{Probing procedure}
\label{sec:procedure}
Our goal is to measure the knowledge stored in \bert{}'s parametric memory via knowledge probes. 
LAMA probes rely on the MLM decoding head to complete cloze statement tasks. Note that this decoder is only trained for the mask token embedding of the final layer and is unsuitable if we want to probe all layers of \bert{}. 
To overcome this, we train an individual decoding head for each layer of a \bert{} model under investigation.

\mpara{Training}: The new decoding head for each layer are trained the same way as in \bert{}'s standard pre-training, by using MLM. 
We also used Wikipedia (WikiText-2 data) -- sampling passages at random and then randomly masking 15\% of the tokens in each.
Our decoding head uses the same architecture as proposed by~\citet{devlin2018bert} -- a fully connected layer with GELU activation and layer norm (epsilon of 1e-12) resulting in a new 768 dimensional embedding. This embedding is then fed to a linear layer with softmax activation to output a probability distribution over the 30K vocabulary terms. In total, the decoding head possesses $\sim$24M parameters. We froze \bert{}'s parameters and trained only the decoding head for every layer using the same training data. We initialized the new decoding heads with the parameters of the pre-trained decoding head and then fine-tuned it. Our experiments with random initialization yielded no significant difference in performance for the lower and middle and worse performance on the last few layers, but resulted in longer training time. We used a batch size of 8 and trained until validation loss was minimized using the AdamW optimizer~\cite{Loshchilov2019DecoupledWD}. With the new decoding heads, the LAMA probes can be applied to every layer. 



\mpara{Measuring knowledge}: We convert the probability distribution output of the decoding head to a ranking with the most probable token at rank 1. The amount of knowledge stored at each layer is measured by precision at rank 1 (P@1 for short). We use P@1 as the main metric in all our experiments. Since rank depth of 1 is a strict metric, we also measured P@10 and P@100. We found the trends to be similar across varying rank depths. For completeness, results for P@10 and P@100 can be found in the Appendix. Additionally, we measure the total amount of knowledge contained in \bert{} by 
\begin{equation}
    \mathcal{P}@1 = max( \{ P^{l}@1  | \,\, \forall l \in L  \} )
\end{equation}

where $L$ is the set of all layers and $P^{l}@1$ is the P@1 for a given layer $l$. In our experiments $|L|=12$. This metric allows us to consider knowledge captured at all layers of \bert{}, not just a specific layer. If knowledge is always best captured at one specific layer $l$ then $\mathcal{P}$@1 = $P^{l}@1$. If the last layer always contains the most information then total knowledge is equal to the knowledge stored in the last layer.  

\mpara{Caveats of probing with cloze statements}: Note that \bert{}, \mlmmarco{}, and \mlmsquad{} are trained for the task of masked word prediction which is exactly the same task as our probes. The last layers of \bert{} have shown to contain mostly task-specific knowledge -- how to predict the masked word in this case~\cite{kovaleva-etal-2019-revealing}. Hence, good performance in our probes at the last layers for MLM models can be partially attributed to task-based knowledge.

\section{Results}
\label{sec:results}

In contrast to existing work, we want to analyze relational knowledge across layers to measure the total knowledge contained in \bert{} and observe the evolution of relational knowledge through the layers. 

\subsection{Intermediate layers matter}
\label{sec:intermediate}

The first question we tackle is -- Does knowledge reside strictly in the last layer of \bert{}? 


\begin{figure*}[h]
    \centering
  \includegraphics[width=0.99\textwidth]{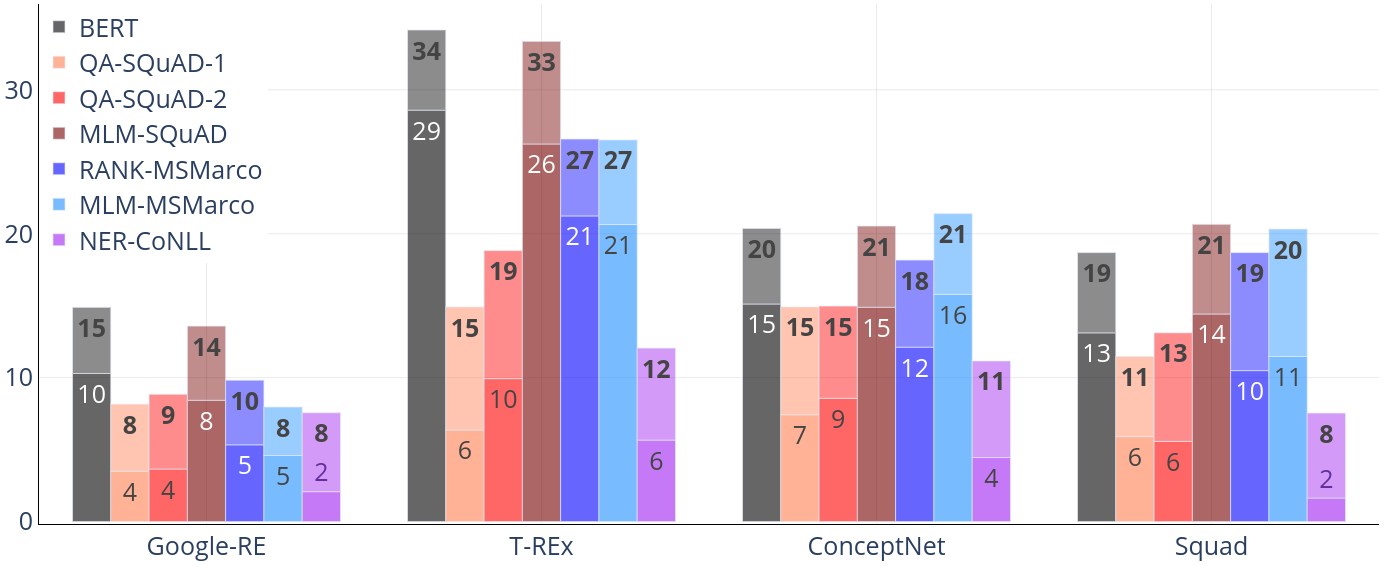}
    \caption{$\mathcal{P}$@1 (upper value) vs last layer P@1 (lower value) for all models for each LAMA probe. Values are averaged over all relations in a dataset and given in percent. \bert{}'s P@1 of 10\% in the last layer (\greprobe{}) means that it answered 10\% of the cloze-questions correctly.}
    \label{fig:last_layer_vs_union}
\end{figure*}

Figure~\ref{fig:last_layer_vs_union} compares the fraction of correct predictions in the last layer as against all the correct predictions computed at any intermediate layer in terms of $\mathcal{P}@1$.
It is immediately evident that a significant amount of knowledge is stored in the intermediate layers.
While the last layer does contain a reasonable amount of knowledge, a considerable proportion of relations seem to be forgotten and the \textbf{intermediate layers contain relational knowledge that is absent in the final layer.}
Specifically, 18\% for \trexprobe{} and 33\% approximately for the others are forgotten by \bert{}'s last layer. 
For instance, the answer to \texttt{Rocky Balboa was born in [MASK]} is correctly predicted as \texttt{Philadelphia} by Layer 10 whereas the rank of \texttt{Philadelphia} in the last layer drops to $26$ for \bert{}.

The intermediary layers also matter for fine-tuned models. Models with high $\mathcal{P}@1$ tend to have a smaller fraction of knowledge stored in the intermediate layers -- 20\% for \rankmarco{} on \trexprobe{}. In other cases, the amount of knowledge lost in the final layer is more drastic -- $3\times$ for \qasquadbig{} on \greprobe{}.

\begin{table}[]
\centering
\small
\begin{tabular}{lccc}
\hline
\textbf{Models}                      & \textbf{P@1} & \textbf{P@10} & \textbf{P@100} \\ \hline
\multicolumn{1}{l}{\bert{}}         & 0.07        & 0.02          & 0.07          \\
\multicolumn{1}{l}{\qasquad{}}    & 0.43        & 0.38         & 0.38          \\
\multicolumn{1}{l}{\qasquadbig{}}    & 0.17        & 0.19         & 0.17          \\
\multicolumn{1}{l}{\mlmsquad{}}   & 0.12        & 0.07         & 0.07          \\
\multicolumn{1}{l}{\rankmarco{}} & 0.02         & 0.05          & 0.05           \\
\multicolumn{1}{l}{\mlmmarco{}}     & 0.10        & 0.10         & 0.14          \\
\multicolumn{1}{l}{\ner}  & 0.26        & 0.33         & 0.43          \\ \hline
\end{tabular}
\caption{\label{tab:win_loss} Fraction of relationship types (of the 41 \trexprobe{}) that are best captured at an intermediary layer. Formally, this is the case if mean $P^{12}@1 <$ mean $P^{l}@1$ (p-value $< 0.05$).}
\end{table}

We also measured the fraction of relationship types in \trexprobe{} that are better captured in the intermediary layers (Table~\ref{tab:win_loss}). On average, 7\% of all relation types in \trexprobe{} are forgotten in the last layer for \bert{}. \rankmarco{} forgets the least amount of relation types (2\%) whereas \qasquad{} forgets the most (43\%) in \trexprobe{}, while also being the least knowledgeable (lowest or second-lowest $\mathcal{P}@1$ in all probes). This is further proof of our claim that \bert{}'s overall capacity can be better estimated by probing all layers. Surprisingly, \rankmarco{} is able to consistently store nearly all of its knowledge in the last layer. We postulate that for ranking in particular, relational knowledge is a key aspect of the task specific knowledge commonly found in the last layers. 

\subsection{Relational knowledge evolution}
\label{sec:evolution}

\begin{figure}[ht!]
    \centering
    \includegraphics[width=0.50\textwidth]{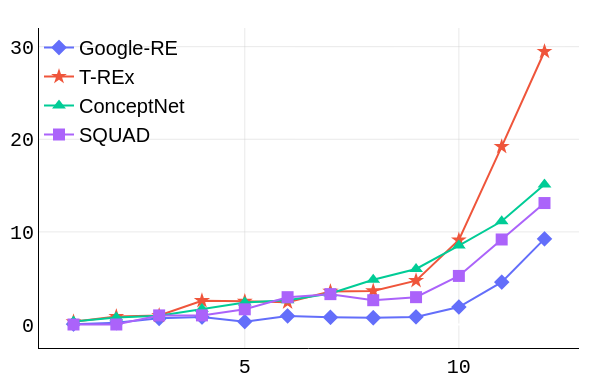}
    \caption{Mean P@1 of \bert{} across all layers.}
    \label{fig:bert_layers}
\end{figure}

Next, we study the evolution of relational knowledge through the \bert{} layers presented in Figure~\ref{fig:bert_layers} that reports P@1 at different layers.

\begin{figure}[ht]
    \centering
    \includegraphics[width=0.5\textwidth]{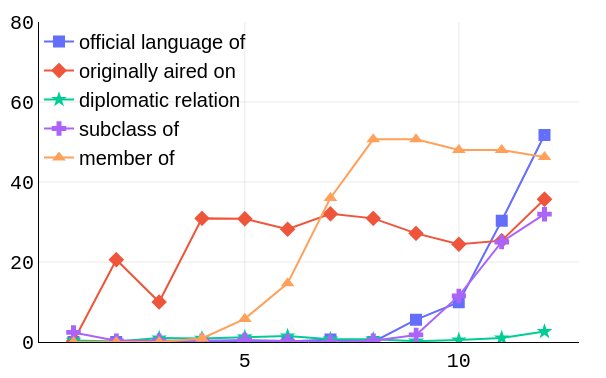}
    \caption{P@1 across all layers for \bert{} for select relationship types from \trexprobe{}.}
    \label{fig:trex_all_default}
\end{figure}

We observe that the amount of \textbf{relational knowledge captured increases steadily with each additional layer}.
While some relations are easier to capture early on, we see an almost-exponential growth of relational knowledge after Layer 8.
This indicates that relational knowledge is predominantly stored in the last few layers as against low-level linguistic patterns are learned at the lower layers (similar to ~\citet{vanaken_2019}). Interestingly, \citet{tenney2019bert} found that the center of gravity for co-reference resolution in layer 16 of \bert{}-large's 24 layers. We hypothesize that a certain proficiency in matching different entity mentions is necessary for attributing facts to an entity. 
In Figure~\ref{fig:trex_all_default} we inspect relationship types that show uncharacteristic growth or loss in \trexprobe{}. 

While \texttt{member of} is forgotten in the last layers, the relation \texttt{diplomatic relation} is never learned at all, and  \texttt{official language of} is only identifiable in the last two layers. Note that the majority of relations follow the nearly exponential growth curve of the mean performance in Figure~\ref{fig:bert_layers} (see line \trexprobe{}). From our calculations, nearly 15\% of relationship types double in mean P@1 at the last layer.

We now analyze evolution in fine-tuned models to understand the impact of fine-tuning on the knowledge contained through the layers.
There are two effects at play once \bert{} is fine-tuned. 
First, during fine-tuning \bert{} observes additional task-specific data and hence has either opportunity to monotonically increase its relational knowledge or replace relational knowledge with more task-specific information. Second, the task-specific loss function might be misaligned with the MLM probing task. 
This means that fine-tuning might result in difficulties in retrieving the actual knowledge using the MLM head.
In the following, we first look at the overall results and then focus on specific effects thereafter.

\begin{figure}[ht!]
    \centering
    \includegraphics[width=0.5\textwidth]{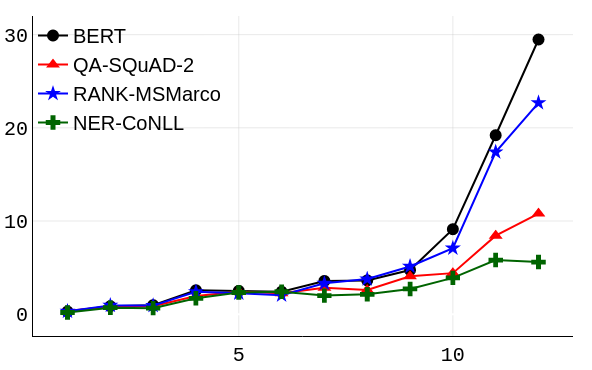}
    \caption{Knowledge contained per layer measured in terms of P@1 on \trexprobe{}.}
    \label{fig:layerwise_mixed}
\end{figure}

Figure~\ref{fig:layerwise_mixed} shows the evolution of knowledge in three fine-tuned models when compared to \bert{}. 

All models possess nearly the same amount of knowledge until layer 6 but then start to grow at different rates. Most surprisingly, \rankmarco{}'s evolution is closest to \bert{} whereas the other models forget information rapidly. With previous studies indicating that the last layers make way for task-specific knowledge
~\cite{kovaleva-etal-2019-revealing}, the ranking model can retain a larger amount of knowledge when compared to other fine-tuning tasks in our experiments.

These results raise the question: Is \rankmarco{} able to retain more knowledge because MSMarco is a bigger dataset or is it because the ranking objective is better suited to knowledge retention as compared to QA, MLM or NER?



\subsection{Effect of fine-tuning objectives}
\label{sec:finetuning}

\begin{figure}
\centering
\begin{tabular}{cc}
  \includegraphics[width=50mm]{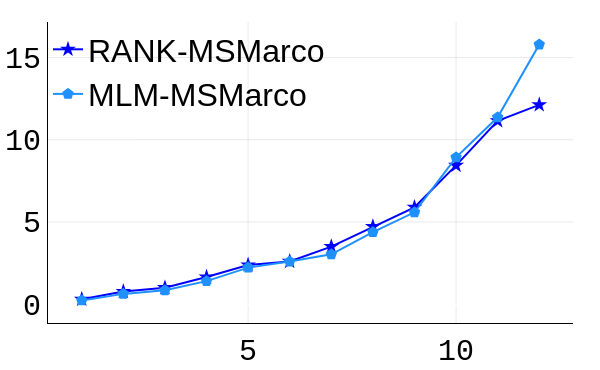} &   \includegraphics[width=50mm]{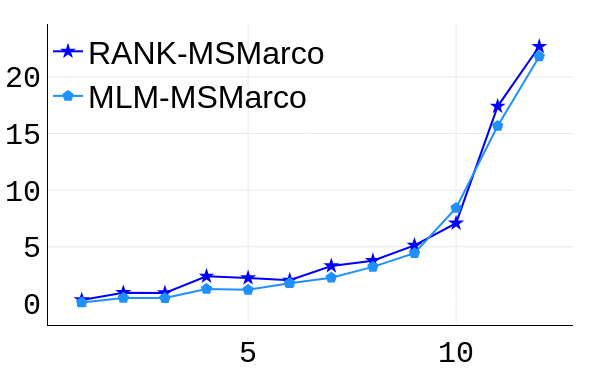} \\
(a) ConceptNet & (b) T-REx \\[4pt]
 \includegraphics[width=50mm]{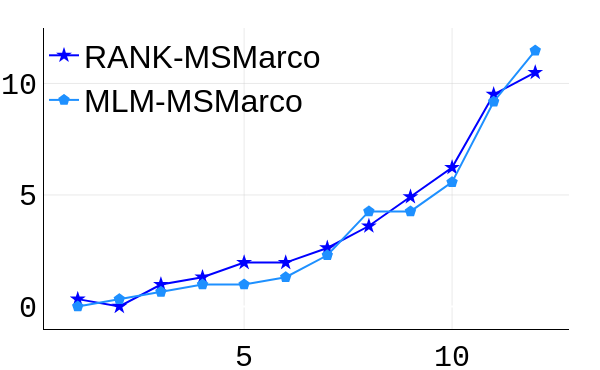} &   \includegraphics[width=50mm]{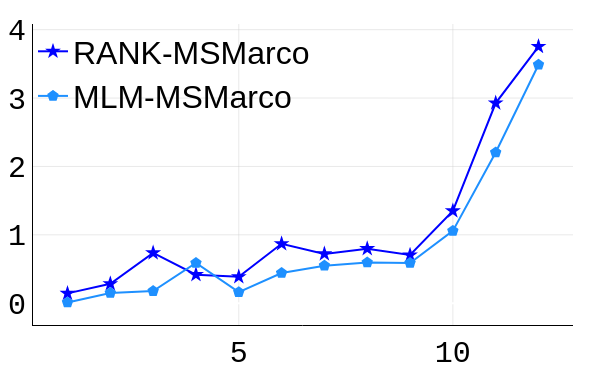} \\
(c) Squad & (d) Google-RE \\[4pt]
\end{tabular}
\caption{Effect of Fine-Tuning Objective on fixed size data: MSMarco.}
\label{fig:finetuning_msm}
\end{figure}

For the first experiment, we study the effect of the fine-tuning objective. As mentioned earlier, it is possible that the task objective function is misaligned with the probing procedure.
Hence, we conducted two experiments where we fixed the dataset and compared the MLM objective (\mlmmarco{}) vs. the ranking objective \rankmarco{} and \mlmsquad{} vs. the span prediction objective (\qasquadbig{}). Figure~\ref{fig:finetuning_msm} shows the evolution of knowledge captured for \mlmmarco{} vs. \rankmarco{}. 

We observe that \rankmarco{} performs quite similar to \mlmmarco{} across all probes and layers. Although \mlmmarco{} has the same training objective as the probe, the ranking model can retain nearly the same amount of knowledge. We hypothesize that this is because the downstream fine-tuning task is sensitive to relational information.
Specifically, ranking passages for open-domain QA is a task that relies heavily on identifying pieces of knowledge that are strongly related -- For example, given the query: \textit{How do you mow the lawn?}, \rankmarco{} must effectively identify concepts and relations in candidate passages that are related to lawn mowing (like types of grass and lawnmowers) to estimate relevance. 

\begin{figure}
\centering
\begin{tabular}{cc}
  \includegraphics[width=50mm]{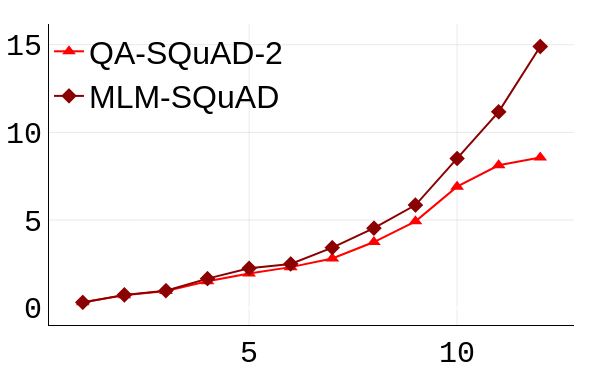} & \includegraphics[width=50mm]{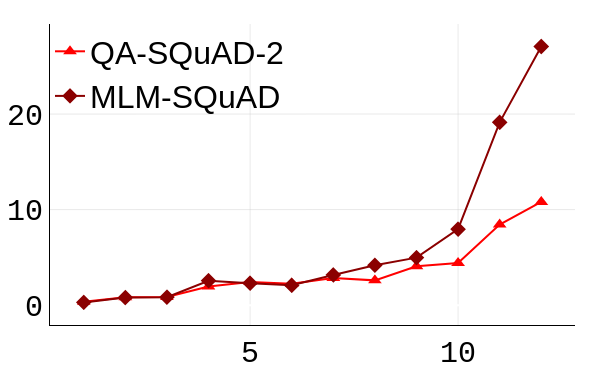} \\
(a) ConceptNet & (b) T-REx \\[4pt]
 \includegraphics[width=50mm]{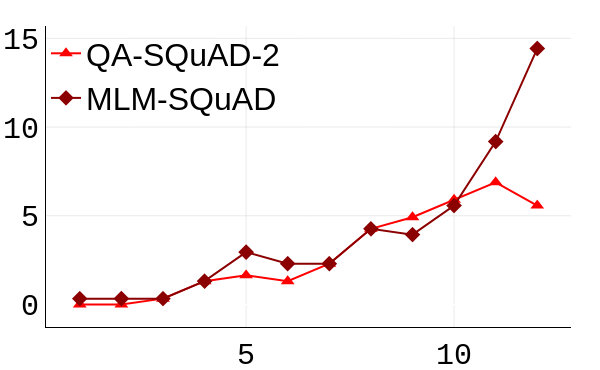} &   \includegraphics[width=50mm]{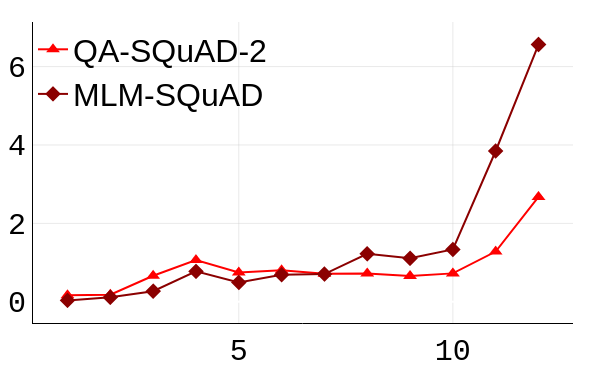} \\
(c) Squad & (d) Google-RE \\[4pt]
\end{tabular}
\caption{Effect of Fine-Tuning Objective on fixed size data: SQUAD.}
\label{fig:finetuning_squad}
\end{figure}

Reading comprehension or span prediction (QA) however seems to be a less knowledge-intensive task both in terms of total knowledge and at the last layer (Figure~\ref{fig:last_layer_vs_union}). In Figure~\ref{fig:finetuning_squad} we see that the final layers are impacted the most by fine-tuning on question answering. Intuitively, span prediction appears to be the tasks that requires less relational knowledge to be remembered as models only need to select the right answer from a given context. From Table~\ref{tab:win_loss} we observe that \mlmsquad{} has a lower amount of relations better captures at intermediary layers (12\% vs 17\%), with \qasquadbig{} seemingly forgoing relational knowledge for span prediction task knowledge. It is hard to compare \qasquadbig{} and \rankmarco{} from this experiment as both were trained on different datasets. In the next section, we study the effect of fine-tuning data on the amount of knowledge captured in parametric memory.

\subsection{Effect of fine-tuning data}
\label{sec:data_size}

To isolate the effect of the fine-tuning datasets, we fix the fine-tuning objective. 
We experimented with the MLM and the QA span prediction objective. 
For MLM, we used models trained on fine-tuning task data of varying size -- \bert{}, \mlmmarco{} ($\sim$ 8.8 million unique passages) and \mlmsquad{} ($\sim$ 500+ unique articles). For the QA objective, we experimented with \qasquad{} and \qasquadbig{} which utilize the same dataset of passages but \qasquadbig{} is trained on 50K extra unanswerable questions.


\begin{figure}
\centering
\begin{tabular}{cc}
  \includegraphics[width=50mm]{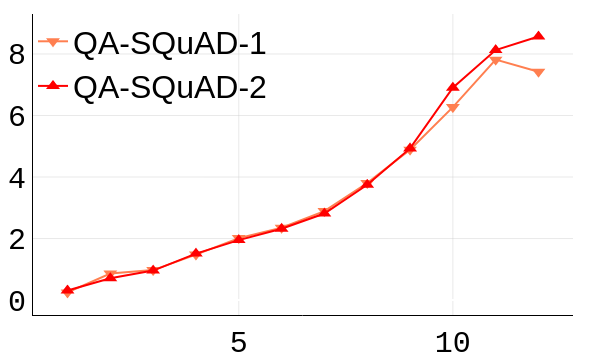} &   \includegraphics[width=50mm]{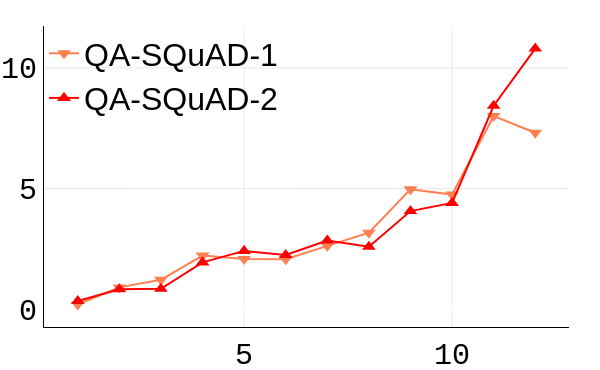} \\
(a) ConceptNet & (b) T-REx \\[4pt]
 \includegraphics[width=50mm]{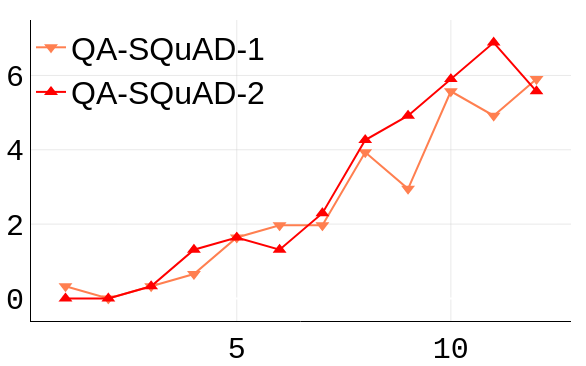} &   \includegraphics[width=50mm]{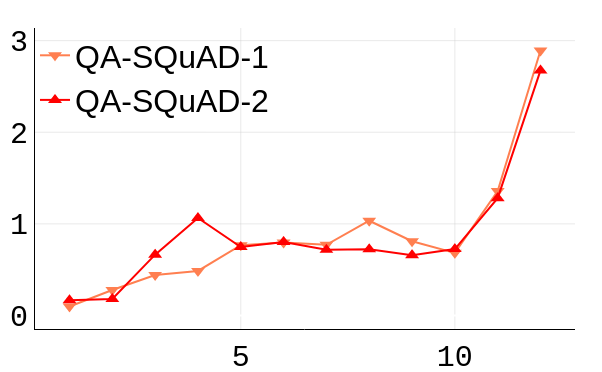} \\
(c) Squad & (d) Google-RE \\[4pt]
\end{tabular}
\caption{Effect of dataset size. Mean P@1 across layers for \qasquad{} and \qasquadbig{}. }
\label{fig:dataset_qa_squad}
\end{figure}

Considering the QA span prediction objective, we first see that the total amount of knowledge stored ($\mathcal{P}@1$) in \qasquadbig{} is higher for 3/4 knowledge probes (from Figure~\ref{fig:last_layer_vs_union}). Figure~\ref{fig:dataset_qa_squad} shows the evolution of knowledge captured for \qasquad{} vs. \qasquadbig{}. \qasquadbig{} captures more knowledge at the last layer in 3/4 probes with both models showing similar knowledge emergence trends. This result hints to the fact that a more difficult task (SQuAD 2) on the same dataset forces BERT to remember more relational knowledge in its final layers as compared to the relatively simpler SQuAD 1. This point is further emphasized in Table~\ref{tab:win_loss}. Only 17\% of relation types are better captured in the intermediary layers of \qasquadbig{} as compared to 43\% for \qasquad{}.


\begin{figure}
\centering
\begin{tabular}{cc}
  \includegraphics[width=50mm]{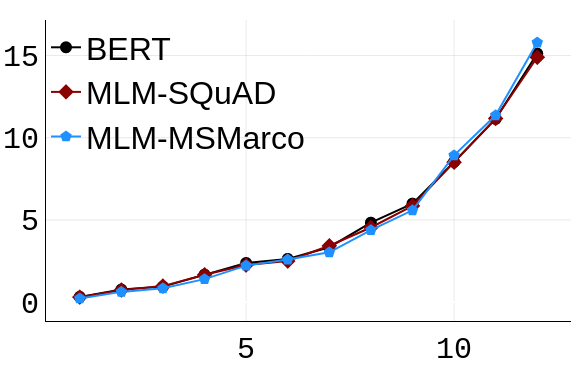} &   \includegraphics[width=50mm]{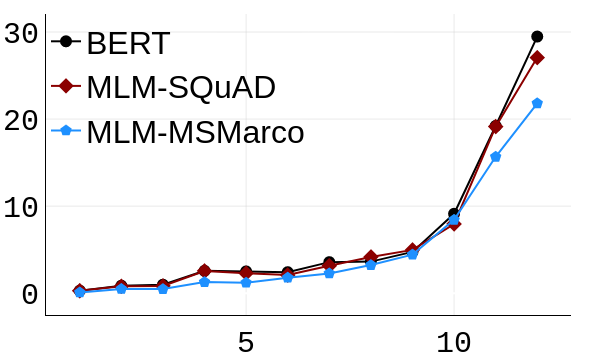} \\
(a) ConceptNet & (b) T-REx \\[4pt]
 \includegraphics[width=50mm]{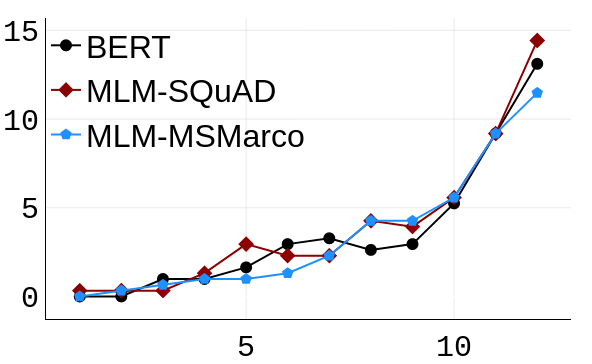} &   \includegraphics[width=50mm]{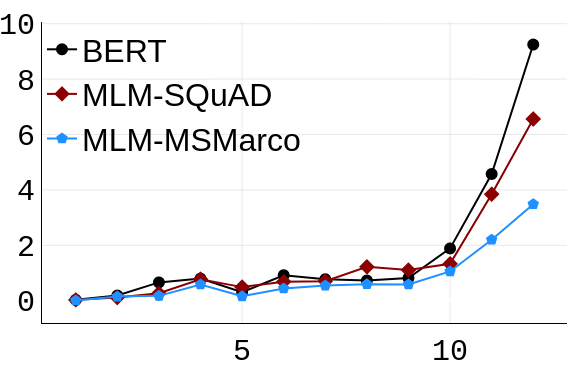} \\
(c) Squad & (d) Google-RE \\[4pt]
\end{tabular}
\caption{Effect of dataset size. Mean P@1 across layers for \bert{}, \mlmmarco{} and \mlmsquad{}.}
\label{fig:dataset_mlm}
\end{figure}

Figure~\ref{fig:dataset_mlm} shows the evolution of knowledge for both MLM models as compared to \bert{}. When being fine-tuning, \bert{} seemingly tends to forget some relational knowledge to accommodate for more domain-specific knowledge. We suspect it forgets certain relations (found in the probe) to make way for other knowledge not detectable by our probes. In the case where the probe is aligned with the fine-tuning data (\squadprobe{}), \mlmsquad{} learns more about its domain and outperforms \bert{}, but only by a small margin ($<5\%$). Even though \mlmmarco{} uses a different dataset it is able to retain a similar level of knowledge in \squadprobe{}. The evolution trends in Figure~\ref{fig:dataset_mlm} further confirm that fine-tuning leads to forgetting mostly in the last layers. Since the fine-tuning objective and probing tasks are aligned, it is more evident in these experiments that relational knowledge is being forgotten or replaced.

When observing $\mathcal{P}@1$ and $P@1$, according to \trexprobe{} and \greprobe{} in particular, \mlmmarco{} forgets a large amount of knowledge but retains common sense knowledge (\concpetprobe{}). \mlmsquad{} contains substantially more knowledge overall according to 2/4 probes and nearly the same in the others as compared to \mlmmarco{}. Seemingly, the amount of knowledge contained in fine-tuned models is not directly correlated with the size of the dataset, on the contrary, we observe that training on the bigger dataset, containing more factual information, leads to more forgetting. There can be several contributing factors to this phenomenon potentially related to the data distribution and alignment of the probes with the fine-tuning data. In the next section, we will further investigate what lead to such a substantial amount of knowledge being dropped by \mlmmarco{} and \mlmsquad{}.

\section{What happens during Fine-tuning?}
\label{sec:fine-tuning-what-happens}

To understand what causes the forgetting observed in Figure~\ref{fig:dataset_mlm}, let us take a step back and revisit what is happening when a model is fine-tuned. As mentioned earlier, the process of fine-tuning usually exposes a model to a new task as well as to new data. One reason for our results might be, that the fine-tuning (QA, ranking, NER) and probing task (MLM) are very different and models suffer from the sequential learning problem. In Section~\ref{sec:sec:catastrophic-forgetting}, we will investigate what effect this alignment between training and probing tasks plays in our results. Other factors that might explain our results are the training \textit{data} being misaligned with the probes (Section~\ref{sec:sec:overlap}) and models dropping facts that are no longer present in the fine-tuning data due to capacity reasons (Section~\ref{sec:sec:capacity}). Nevertheless, exposing models to new data includes the possibility of learning new facts (Section~\ref{sec:sec:learn-forget}). In the following, we will discuss these points one by one.  


\subsection{Catastrophic forgetting}
\label{sec:sec:catastrophic-forgetting}

With the current transformer-based language models, fine-tuning tasks are common down-stream tasks such as ranking, question answering or sentiment analysis, whereas pre-training tasks typically focus on teaching the models general language understanding (e.g., masked language modeling or causal language modeling). Fine-tuning tasks tend to be more specific than the pre-training tasks, hence, requiring a different set of (task) knowledge to effectively solve. As our probing task is masked language modeling, models might lose the ability to perform MLM after being fine-tuned on a different task. 

\begin{table}[h]
\centering
\begin{tabular}{lc}
\hline
\textbf{Model}      & \multicolumn{1}{l}{\textbf{MLM test loss}} \\ \hline
\bert{}                & 2.12                                      \\
\qasquad{}  & 2.93                                     \\
\qasquadbig{}  & 2.78                                      \\
\mlmsquad{} & 1.76                                      \\
\mlmmarco{}  & 1.93                                      \\
\rankmarco{}  & 2.06                                       \\
\ner{} & 3.76                                    \\ \hline
\end{tabular}
\caption{MLM performance of the models after being fine-tuned. Last layer, MLM head is re-trained for all non-MLM models. Evaluated on the Wikitext-2 test set.}
\label{tab:catastrophic_forgetting_mlm}
\end{table}

Table~\ref{tab:catastrophic_forgetting_mlm} shows the ability to do masked language modeling for \bert{} and the fine-tuned models as tested on the Wikitext-2 dataset. Note, that the non-MLM models were tested after re-training the decoding head as described Section~\ref{sec:setup}. Thus, this measures how much MLM ability can be decoded from the fine-tuned model's embeddings. In accordance with our previous probing results, we observe \qasquad{}, \qasquadbig{} and \ner{} to be on the lower end, while \bert{}, \rankmarco{} and the MLM-models perform better. This indicates, that a part of the performance difference is indeed because of different alignments between fine-tuning tasks and masked language modeling. Nevertheless, as \mlmsquad{} and \mlmmarco{} both have a lower MLM loss than \bert{} while generally performing worse on the probes. Catastrophic forgetting can not explain the different amounts of factual knowledge in these models.

\subsection{Overlap between training and probing data}
\label{sec:sec:overlap}

Another reason why \mlmmarco{} performed worse than \mlmsquad{}, despite MSMarco containing more facts, might be, that training and probing data are not aligned. That is, the model learns facts from MSMarco, that are not queried for in our probe set, which is mostly derived from Wikipedia. 
To understand how the probing facts are represented in SQuAD and MSMarco, we devise a lightweight matching methodology: 
First, we compute an inverted index over the training datasets, regarding question + context (SQuAD) and query + passage (MSMarco) as individual documents. Second, we check for each probe (containing subject, relation, object) if one document in the inverted index contains both the subject and the object. While this matching is not perfect, we deem this to be sufficient to give us some insight on what proportion of facts is covered in the training data.
Results of this alignment between training and probing data can be found in Table~\ref{tab:info_overlap}.

\begin{table*}[h]
\centering
\begin{tabular}{lccc}
\hline
\textbf{Training dataset} & \textbf{\greprobe{}} & \textbf{\trexprobe{}}  & \textbf{Information density}           \\ \hline
SQuAD            & 0.51\%    & 11.1\% & 3393 facts in 100k passages   \\
MSMarco          & 1.9\%       & 38\%   & 12.621 facts in 8.8M passages \\ \hline
\end{tabular}
\caption{Estimated overlap between the fine-tuning datasets (SQuAD, MSMarco) and the probe sets (\greprobe{}, \trexprobe{}).}
\label{tab:info_overlap}
\end{table*}

Surprisingly, we observe only a low amount of \greprobe{} facts covered in both probing sets, while both dataset contain a significantly larger proportion of \trexprobe{} facts. Note, the English Wikipedia contains over 6M articles and our probing facts span $\sim$ 52k facts, resulting in many facts not being covered by the probes. Also, while we find MSMarco to contain 3.5x more of our facts than SQuAD, this is surprisingly little as MSMarco also spans 88x more passages. The quotient of the number of facts covered in the dataset relative to the number of passages will be from here on referred to as the information density. This notion will give us a feel for how sparse or dense our probing facts are covered by the dataset. We conclude that both probing sets must contain a high amount of facts that are not covered by our probe sets. When restricting the capacity to retain information, under-represented classes are dropped \citep{Hooker2020WhatDC}). Given the low frequency of facts in text (compared to stop words) and the low information density of probing facts observed in MSMarco and SQuAD, we question if the lower probing results for \mlmmarco{} and \mlmsquad{} are due to models replacing unused knowledge with other facts.

\subsection{Capacity}
\label{sec:sec:capacity}

To understand how capacity affects the models retention of facts, we deploy the probe tasks used throughout this paper. These are very useful as they contain both sentences expressing only the fact as well as the evidence sentences from which the facts where extracted from. In this experiment, we use the facts as training as well as test data. This allows understanding how much and which information is actually learned and what is forgotten. This time, we do not shuffle our training data but train \bert{} sequentially on the probe sets in order of \greprobe{}, \trexprobe{}, \concpetprobe{} and \squadprobe{}. We now train \bert{} for different numbers of epochs on the templates and evidences (Table~\ref{tab:facts_template_evidence}). The results are shown in Table~\ref{tab:capacity}.

\begin{table}[h]
\begin{tabularx}{\textwidth}{lX}
\hline
\textbf{Modality} & \textbf{Representation}                                                                                                    \\ \hline
Fact-triple              & (\textit{Albert Einstein}, born-in, \textit{Ulm})                                                                                            \\
Template          & \textit{Albert Einstein} was born in \textit{Ulm}.                                                                                            \\
Evidence          & \textit{Albert Einstein} was born in \textit{Ulm}, in the \textit{Kingdom of Württemberg} in the \textit{German Empire}, on \textit{14 March 1879} into a family of secular \textit{Ashkenazi Jews}. His parents were \textit{Hermann Einstein}, a salesman and engineer, and \textit{Pauline Koch}... \\ \hline
\end{tabularx}
\caption{Overview of different representations of a fact. Triples are derived from knowledge bases, templates are sentences expressing only the fact and evidences are sentences these facts were derived from.}
\label{tab:facts_template_evidence}
\end{table}

\begin{table*}[h]
\centering
\begin{tabular}{lccccc}
\hline
\textbf{Model}      & \multicolumn{1}{l}{\textbf{Google-RE}} & \multicolumn{1}{l}{\textbf{T-REx}} & \multicolumn{1}{l}{\textbf{ConceptNet}} & \multicolumn{1}{l}{\textbf{Squad}} & \multicolumn{1}{l}{\textbf{MLM loss}} \\ \hline
BERT                & 0.10                                   & 0.29                               & 0.15                                    & 0.13                               & 2.115                                      \\ \hline
Templates-object-1  & 0.11                                   & 0.44                               & 0.36                                    & 0.36                               & 2.878 (+36\%)                              \\
Templates-object-2  & 0.12                                   & 0.48                               & 0.53                                    & 0.75                               & 3.453 (+63\%)                              \\
Templates-object-10 & 0.96                                   & 0.87                               & 0.88                                    & 0.99                               & 5.419 (+156\%)                             \\ \hline
Evidences-object-1  & 0.10                                   & 0.34                               & 0.39                                    & 0.43                               & 3.156 (+49\%)                              \\
Evidences-object-2  & 0.10                                   & 0.30                               & 0.54                                    & 0.86                               & 4.18 (+98\%)                               \\
Evidences-object-10 & 0.17                                   & 0.36                               & 0.86                                    & 0.99                               & 6.041 (+184\%)                             \\ \hline
\end{tabular}
\caption{Capacity results for training 1,2,10 epochs on the probing data. The training dataset is not shuffled and is in the same order as the table: \greprobe{}, \trexprobe{}, \concpetprobe{}, \squadprobe{}. For masked language modeling, we either use the templates or the evidence sentences from the data source (see Table~\ref{tab:facts_template_evidence}) and mask the objects.}
\label{tab:capacity}
\end{table*}

When training for one epoch on the templates (Templates-object-1), we can see that more of the facts are remembered that were exposed last to the model. Actually, the amount of facts retained increases monotonically. This result suggests two things: Seeing facts once, does not result in facts being perfectly memorized. This is in line with \citet{kassner-etal-2020-symbolic-reasoners}, who found that \bert{} requires seeing facts 15 times to remember them perfectly. Furthermore, it also indicates that most recent knowledge is remembered the best, just as humans do. Yet, even one epoch on the probe sets results in a drop in general MLM performance (+36\% loss), signalling that MLM knowledge had to be dropped to make room for the factual knowledge. 
For training 2 epochs, we observe the same trends. Nevertheless, when training 10 epochs on the probe sets, most of the facts are actually retained by \bert{} with a significant drop in MLM performance (+156\%). 
Training \bert{} on the evidence sentences introduces more facts as distractors into the training process. Overall, this results in increased degradation of MLM performance (+30\% higher compared to when using templates). In contrast to training with template sentences, when training 10 epochs on the evidence sentences, we do not see all facts being remembered close to perfectly. This hints at capacity constraints limiting the amount of facts that can be remembered without dropping too much MLM knowledge. \todo{Note that masking objects is better at learning facts than masking random words? If yes, where? }

\subsection{The circle of learning and forgetting}
\label{sec:sec:learn-forget}
In this section we want to review what happens during fine-tuning on a fact-by-fact instead of a probing set level. A probing set based analysis only shows that most fine-tuned models perform worse than \bert{}. Yet, some questions remain: is knowledge only forgotten or is some knowledge dropped and new knowledge learned? How can we specify what forgotten and learned mean? This is what we intend to answer in this section. 

\begin{figure}[h]
    \centering
    \includegraphics[width=0.5\textwidth]{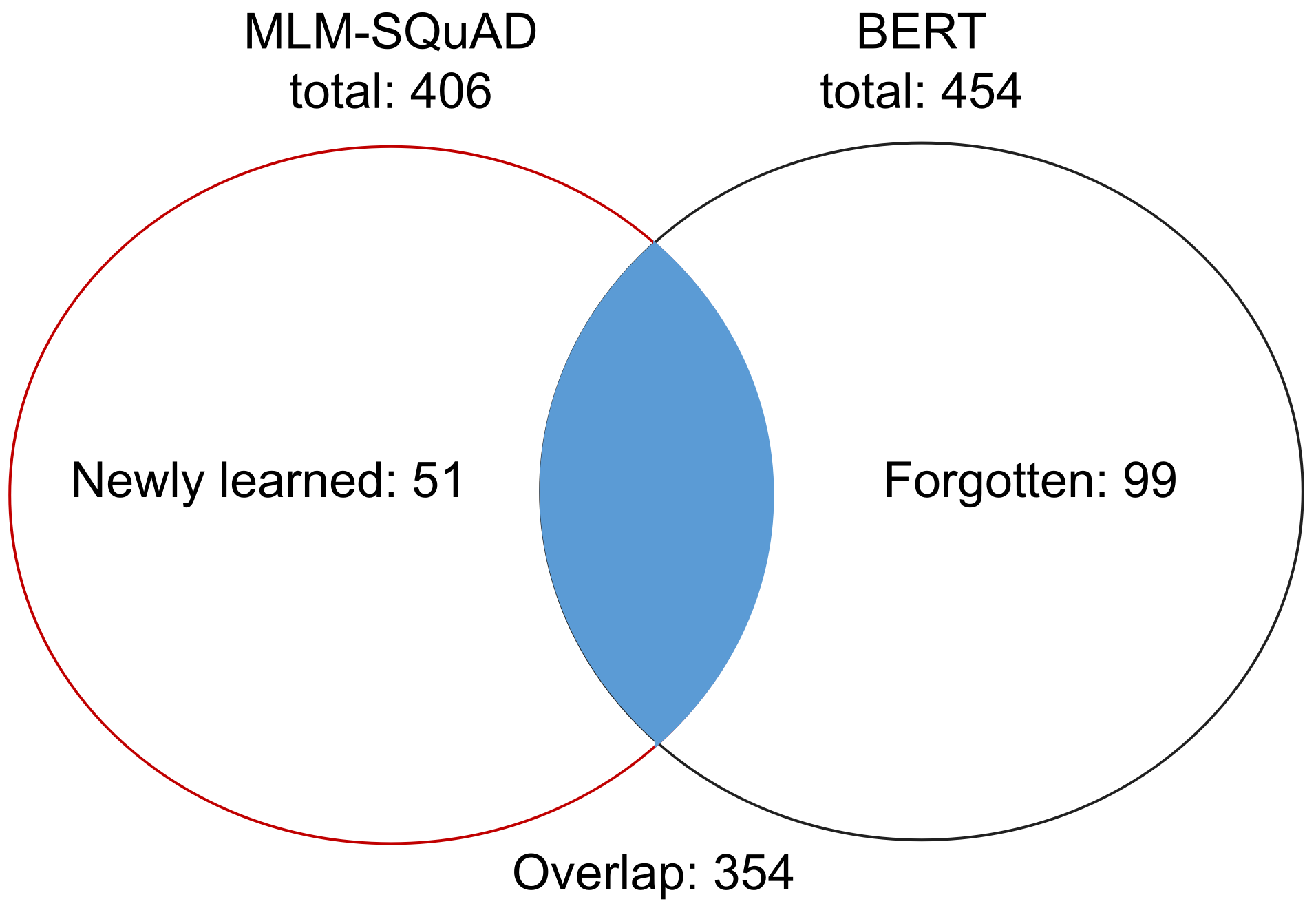}
    \caption{Definition of learned and forgotten.}
    \label{fig:venn}
\end{figure}

To understand what is dropped and acquired by our models, we again consider the entirety of the models and not only the last layers. We define that a model learned something by fine-tuning if some factual probe can be answered by some layer of a fine-tuned model if no layer of \bert{} can answer the probe: that is, the knowledge has not been available anywhere in \bert{}. In that case, the knowledge is indeed newly acquired. Conversely, we define a fact to be forgotten, if a fine-tuned model is no longer able to answer a factual probe at any layer that \bert{} was able to answer correctly (at one of its layer). Visually, what is learned and forgotten can be interpreted as a Venn-diagram (Figure~\ref{fig:venn}), where we have some overlap in the knowledge of models and some knowledge that only one of the models have. 


\begin{equation}
    learned = (\textit{MLM-SQuAD} \cap \textit{Overlap}) / \bert{} = 51/454 \approx 11\%
\end{equation}


\begin{equation}
    forgotten = (\bert{} \cap \textit{Overlap}) / \bert{} = 99/454 \approx 22\%
\end{equation}

\todo{Only keep one of both? Venn or formula?}

An overview of how much knowledge is actually learned and forgotten by the fine-tuned models can be found in Table~\ref{tab:learned_forgotten}.

\begin{table}[h]
\centering
\begin{tabular}{lcccccccccc}
\hline
\textbf{Models}                   & \multicolumn{10}{c}{\textbf{Probing set}}                                                                                                                                                                                                  \\ \hline
\multicolumn{1}{l|}{}             & \multicolumn{2}{c|}{Google-RE}                   & \multicolumn{2}{c|}{T-REx}                       & \multicolumn{2}{c|}{ConceptNet}                  & \multicolumn{2}{c|}{Squad}                       & \multicolumn{2}{c}{Average}    \\ \cline{2-11} 
\multicolumn{1}{l|}{}             & \multicolumn{1}{c|}{-} & \multicolumn{1}{c|}{+}  & \multicolumn{1}{c|}{-} & \multicolumn{1}{c|}{+}  & \multicolumn{1}{c|}{-} & \multicolumn{1}{c|}{+}  & \multicolumn{1}{c|}{-} & \multicolumn{1}{c|}{+}  & \multicolumn{1}{c|}{-} & +     \\ \cline{2-11} 
\multicolumn{1}{l|}{QA-SQuAD-1}   & 68                     & \multicolumn{1}{c|}{28} & 63                     & \multicolumn{1}{c|}{11} & 40                     & \multicolumn{1}{c|}{13} & 49                     & \multicolumn{1}{c|}{11} & 55                     & 15.75 \\
\multicolumn{1}{l|}{QA-SQuAD-2}   & 70                     & \multicolumn{1}{c|}{31} & 56                     & \multicolumn{1}{c|}{13} & 40                     & \multicolumn{1}{c|}{13} & 47                     & \multicolumn{1}{c|}{18} & 53.25                  & 18.75 \\
\multicolumn{1}{l|}{MLM-SQuAD}    & 45                     & \multicolumn{1}{c|}{27} & 21                     & \multicolumn{1}{c|}{19} & 15                     & \multicolumn{1}{c|}{16} & 14                     & \multicolumn{1}{c|}{25} & 23.75                  & 21.75 \\
\multicolumn{1}{l|}{MLM-MSMarco}  & 72                     & \multicolumn{1}{c|}{28} & 43                     & \multicolumn{1}{c|}{29} & 26                     & \multicolumn{1}{c|}{31} & 28                     & \multicolumn{1}{c|}{37} & 42.25                  & 31.25 \\
\multicolumn{1}{l|}{RANK-MSMarco} & 58                     & \multicolumn{1}{c|}{14} & 33                     & \multicolumn{1}{c|}{14} & 23                     & \multicolumn{1}{c|}{13} & 25                     & \multicolumn{1}{c|}{25} & 34.75                  & 16.5  \\ \hline
\end{tabular}
\caption{Proportion of (the pre-trained) \bert{}'s knowledge that is forgotten (-) by fine-tuning and the amount that is newly acquired (+). If a model learned 10\% it means that during its training, it acquired 10\% additional knowledge that was not in \bert{} before fine-tuning.}
\label{tab:learned_forgotten}
\end{table}

First of all, we find all models to learn a significant amount of knowledge from fine-tuning (11-37\% compared to \bert{}). In accordance with our previous findings that the QA models performing sub-par, we observe \qasquad{} and \qasquadbig{} to forget the most facts. When the fine-tuning objective is MLM (\mlmsquad{}), we observe a slightly smaller amount of facts being forgotten and a higher amount of new facts being acquired (22\% on average). Yet, \qasquad{} and \qasquadbig{} also learn a decent amount of new facts (16\% and 19\% respectively). In comparing the QA models, we again see the model trained on the harder task outperforming \qasquad{}. Turning to the two MLM models \mlmmarco{} and \mlmsquad{}, the amount of facts being forgotten is consistently higher for \mlmmarco{}, which was trained on the vastly bigger dataset. This can be the result of limited knowledge capacity forcing \mlmmarco{} to drop facts for new facts being learned. Accordingly, \mlmmarco{} outperforms \mlmsquad{} when it comes to new knowledge being acquired (31\%). Lastly, \rankmarco{} shows less knowledge being forgotten but also less new facts being learned compared to \mlmmarco{}.

\section{Discussion and Conclusion}
\label{sec:discussion_conclusion}

\todo{Where do I want to mention this? Own subsection "On MLM and factual knowledge? There, we could also mention that MLM learn more facts when masking objects instead of random tokens.
We also note that remembering factual information is not the goal of masked language modeling. Language models are pre-trained on MLM and related tasks to gain a general understanding of language. When randomly masking out words in text, only few predictions require remembering factual information: most can be either guessed from the context or by learning the general linguistic patterns that our languages build on.}

In this paper, we introduce a framework to probe all layers of \bert{} for knowledge.
We experimented on a variety of probes and fine-tuning tasks and found that \bert{} contains more knowledge than was reported earlier. Our experiments shed light on the hidden knowledge stored in \bert{}'s parametric memory and also some important implications to model building. 
Since intermediate layers contain knowledge that is forgotten by the final layers to make way for task-specific knowledge, our probing procedure can more accurately characterize the knowledge stored.

We show that factual knowledge, like syntactic and semantic patterns, is also replaced at the last layers when models are fine-tuned. However, the last layer can also make way for more domain specific knowledge when the fine-tuning objective is the same as the pre-training objective (MLM) as observed in \squadprobe{}. 
The forgetting that we observe after fine-tuning can be attributed to multiple effects: some tasks seem to be not well aligned to MLM (catastrophic forgetting), whereas other models drop factual knowledge because of capacity or the under-representation of probing facts (information density). Hence, forgetting is not mitigated by larger datasets which potentially contain more factual knowledge (\mlmmarco{} $<$ \mlmsquad{} as measured by $\mathcal{P}$@1).

When investigating different training objectives, we find that knowledge-intensive tasks like ranking better mitigate forgetting compared to span prediction. Although the fine-tuned models generally contained less factual knowledge, with significant (and expected) forgetting in the last layers, \rankmarco{} remembers relatively more relationship types than \bert{} (2\% vs 7\% of \trexprobe{}'s relations being better captured at an intermediary layer, Table~\ref{tab:win_loss}). This result can partially explain findings in \citet{chang2019pre} where they found that pre-training \bert{} with inverse cloze tasks aids it's transferability to a retrieval and ranking setting. Essentially, ranking tasks encourage the retention of factual knowledge (as measured by cloze tasks) since they are seemingly required for reasoning between the relative relevance of documents to a query.
This is also evident when comparing \qasquadbig{} to \rankmarco{}, where the ranking model forgets a smaller amount of facts than the QA model (35\% vs. 53\%). When it comes to acquiring new knowledge from the training process, we find MLM to be the most effective task, whereas we do not see mayor differences between ranking and QA objectives. However, we see potential in either modifying the training procedures to lay more emphasis on learning facts for example by masking entities \citep{sun2019ERNIEER}, or by reducing the capacity load of distractors and low information text (as observed in Table~\ref{tab:capacity}). The latter could be done with either adjusting the attention modules to give longer expiration times to factual knowledge \citep{Sukhbaatar2021NotAM} or reducing the load by jointly masking highly correlated n-grams \citep{Levine2020PMIMaskingPM}. This is to be done in future work. 

With this initial work, we provide a first starting point into understanding how different training objectives effect factual knowledge in language models.
Our results have direct implications on the use of \bert{} as a knowledge base. By effectively choosing layers to query and adopting early exiting strategies \citep{Xin2020DeeBERTDE, xin-etal-2021-berxit} knowledge base completion can be improved. The performance of \rankmarco{} also warrants further investigation into ranking models with different training objectives -- pointwise (regression) vs pairwise vs listwise. More knowledge-intensive QA models like answer generation models may also show a similar trend as ranking tasks but require investigation. We also believe that our framework is well suited to studying variants of \bert{}, different pre-training as well as fine-tuning tasks.
We hope this spurs further research into how the parametric memory of language models works and how we can build more knowledgeable models by devising more effective training paradigms.

As concrete applications of our work, we feel that knowledge intensive tasks that involve access to external knowledge will benefit most from our analysis~\citep{petroni2020kilt,zhang21:expred}.
Additionally, Web tasks that rely on triplified knowledge like tags and relations~\citep{holzmann2017exploring} can employ BERT-based models to store relational information without direct need for learning grammatical and fine-grained linguistic knowledge.
Information retrieval tasks like conversational search and Web search can use our observations to complement their ability to provide clarifications~\citep{kiesel2018toward:conv} or  explanations~\citep{singh2018posthoc:secondary,SinghA19:exs} .



\newpage

\bibliography{anthology,emnlp2020}
\bibliographystyle{compling}


\clearpage

\section{Appendix}

    
    
    
    
    

\subsection{Models}
\begin{itemize}
\item \bert: Off the shelf "bert-base-uncased" from the huggingface transformers library \cite{Wolf2019HuggingFacesTS}
\item \qasquad: Both SQuAD QA models are trained with the huggingface question answering training script \footnote{ https://github.com/huggingface/transformers}. This adds a span prediction head to the default \bert, i.e., a linear layer that computes logits for the span start and span end. So for a given question and a context, it classifies the indices in in which the answer starts and ends. As a loss function it uses crossentropy. The model was trained on a single GPU. We used the huggingface default training script and standard parameters: 2 epochs, learning rate 3e-5, batch size 12. The training dataset was SQuAD 1.1~\cite{rajpurkar2016squad} and it achieved an F1 score of 88.5 on the test set. 
\item \qasquadbig: Single GPU, also using huggingface training script with standard parameters. Learning rate was 3e-5, batch size 12, best model after 2 epochs. The training dataset was SQuAD 2~\cite{DBLP:journals/corr/abs-1806-03822} and it achieved an F1 score of 67 on the test set. 
\item \mlmsquad: \bert{} fine-tuned on text from SQUAD using the masked language modeling objective as per~\citet{devlin2018bert}. 15\% of the tokens masked at random. Trained for 4 epochs with LR 5e-5. Single GPU.
\item \rankmarco: Ranking model trained on the MSMarco passage reranking task~\cite{bajaj2016ms}. We used the fine-tuning procedure described in \cite{nogueira2019passage} to obtain a regression model that predicts a relevance score given query and passage. MSMarco, 100k iterations with batch size 128 (on a TPUv3-8).
\item \mlmmarco: \bert{} fine-tuned on the passages from the MSMarco dataset using the masked language modeling objective as per~\cite{devlin2018bert}. 15\% of the tokens masked at random. 3 epochs, batch size 8, LR 5e-5. Single gpu.
\end{itemize}



\subsection{Datasets:}
\begin{itemize}
    \item Wikitext-2: Used for fine-tuning the MLM head. Subset of the Englisch Wikipedia for long term dependency language modeling. 2,088,628 tokens for training, 217.646 for validation, 245.569 for testing. Vocab size: 33,278 out of vocab: 2.6\% of tokens. It can be downloaded from here: https://www.salesforce.com/products/einstein/ai-research/the-wikitext-dependency-language-modeling-dataset/
    \item LAMA probe data: Can be downloaded from their github: https://github.com/facebookresearch/LAMA . Only used for testing. Consists of: Google-RE: 5528 instances over 3 relations. T-REx: 34017 instances over 41 relations. ConceptNet: 12514 instances. This is not grouped into relations. Squad: 305 instances. Context in-sensitive questions rewritten to cloze-statements. No specific relation either.
    \item SQuAD 1.1: Can be downloaded from here: https://rajpurkar.github.io/SQuAD-explorer/ . 100,000+ question answer pairs based on wikipedia articles. Produced by crowdworkers.
    \item SQuAD 2: Can be downloaded from here: https://rajpurkar.github.io/SQuAD-explorer/ . Combines the 100,000+ question answer pairs with 50,000 unanswerable questions. 
    \item MSMARCO: Can be downlaoded from here: https://microsoft.github.io/msmarco/ . For ranking: Dataset for passage reranking was used. Given 1,000 passages, re-rank by relevance. Dataset contains 8,8m passages. For MLM training: Dataset for QA was used. It consists of over 1m queries and the 8,8m passages. Each query has 10 candidate passages. For MLM, we appended the queries with all candidate passages before feeding into BERT.
\end{itemize}

\newpage

\subsection{Knowledge captured in BERT}
\subsubsection{Intermediate layers matter}
\begin{figure*}[ht!]
\centering
\begin{subfigure}{.24\textwidth}
  \centering
  \includegraphics[width=.99\linewidth]{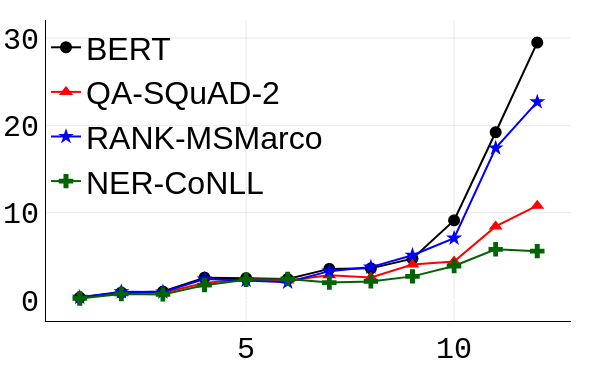}
  \caption{P@1}
\end{subfigure}%
\begin{subfigure}{.24\textwidth}
  \centering
  \includegraphics[width=.99\linewidth]{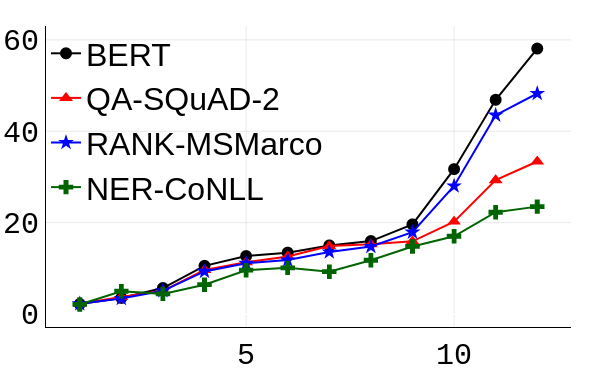}
  \caption{P@10}
\end{subfigure}%
\begin{subfigure}{.24\textwidth}
  \centering
  \includegraphics[width=.99\linewidth]{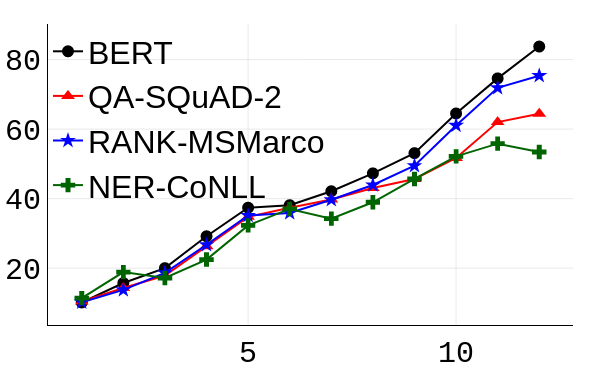}
  \caption{P@100}
\end{subfigure}
\caption{Mean performance in different precisions on \trexprobe{} sets for \bert, \qasquadbig, \rankmarco, \ner.}
\label{fig:layerwise_mixes_additional_precisions}
\end{figure*}

Additional precisions for Figure \ref{fig:layerwise_mixed} can be found in Figure \ref{fig:layerwise_mixes_additional_precisions}.

\subsubsection{Relational knowledge evolution}
\begin{figure*}[ht!]
\centering
\begin{subfigure}{.24\textwidth}
  \centering
  \includegraphics[width=.99\linewidth]{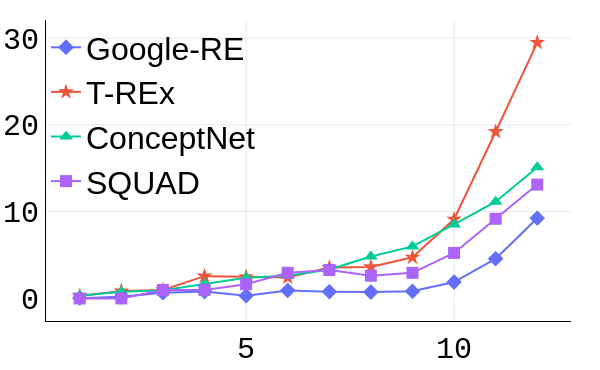}
  \caption{P@1}
\end{subfigure}%
\begin{subfigure}{.24\textwidth}
  \centering
  \includegraphics[width=.99\linewidth]{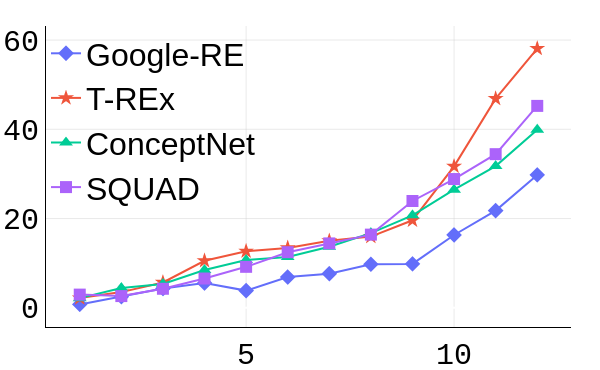}
  \caption{P@10}
\end{subfigure}%
\begin{subfigure}{.24\textwidth}
  \centering
  \includegraphics[width=.99\linewidth]{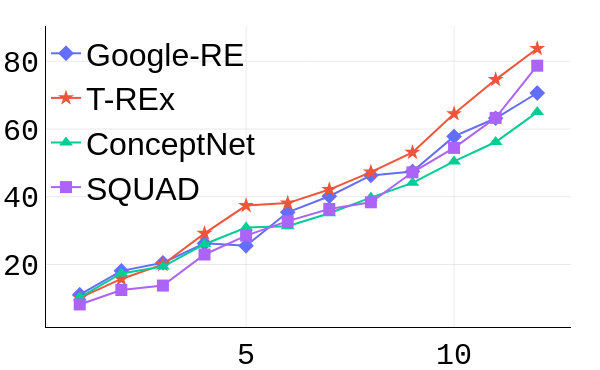}
  \caption{P@100}
\end{subfigure}
\caption{Mean  performance of \bert{} across all layers and probe sets.}
\label{fig:bert_layers_additional_precisions}
\end{figure*}

Additional precisions for Figure \ref{fig:bert_layers} can be found in Figure \ref{fig:bert_layers_additional_precisions}.

\subsubsection{Effect of dataset size}
\begin{figure*}
\begin{subfigure}{.24\textwidth}
  \centering
  \includegraphics[width=.99\linewidth]{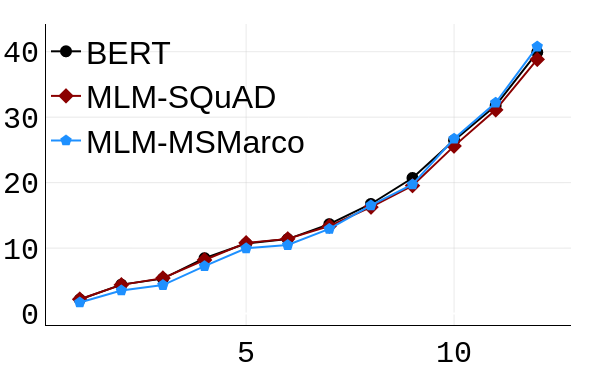}
  \caption{ConceptNet}
\end{subfigure}%
\begin{subfigure}{.24\textwidth}
  \centering
  \includegraphics[width=.99\linewidth]{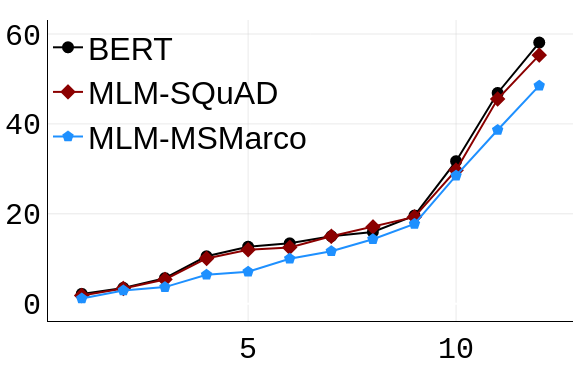}
  \caption{T-REx}
\end{subfigure}%
\begin{subfigure}{.24\textwidth}
  \centering
  \includegraphics[width=.99\linewidth]{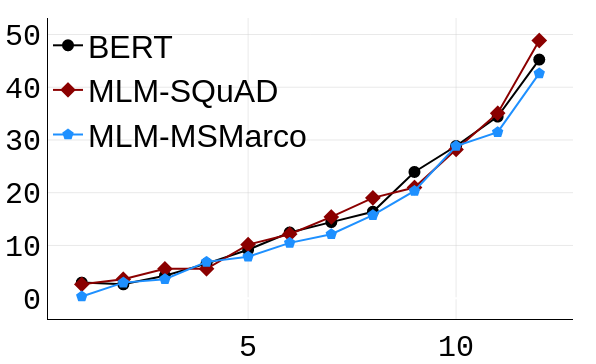}
  \caption{Squad}
\end{subfigure}
\begin{subfigure}{.24\textwidth}
  \centering
  \includegraphics[width=.99\linewidth]{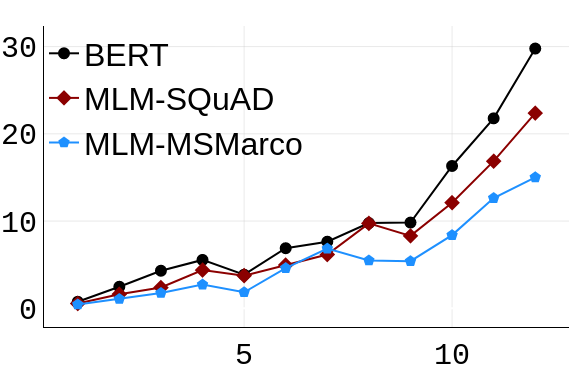}
  \caption{Google-RE}
\end{subfigure}
\caption{Effect of dataset size. Showing P@10}
\label{fig:dataset_mlm_p_10}
\end{figure*}

\begin{figure*}
\begin{subfigure}{.24\textwidth}
  \centering
  \includegraphics[width=.99\linewidth]{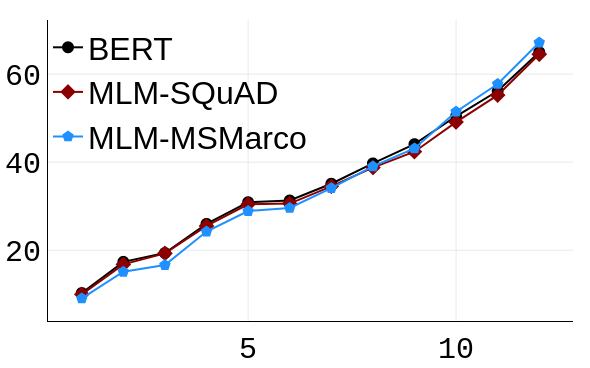}
  \caption{ConceptNet}
\end{subfigure}%
\begin{subfigure}{.24\textwidth}
  \centering
  \includegraphics[width=.99\linewidth]{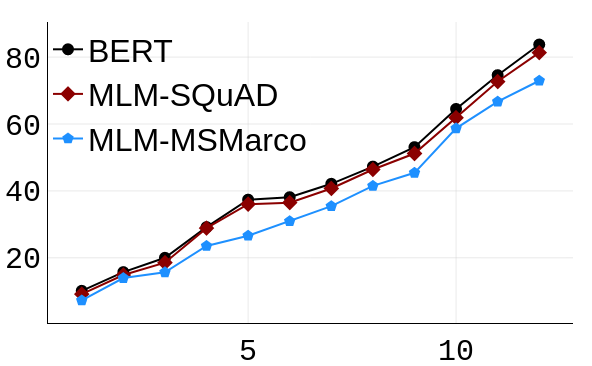}
  \caption{T-REx}
\end{subfigure}%
\begin{subfigure}{.24\textwidth}
  \centering
  \includegraphics[width=.99\linewidth]{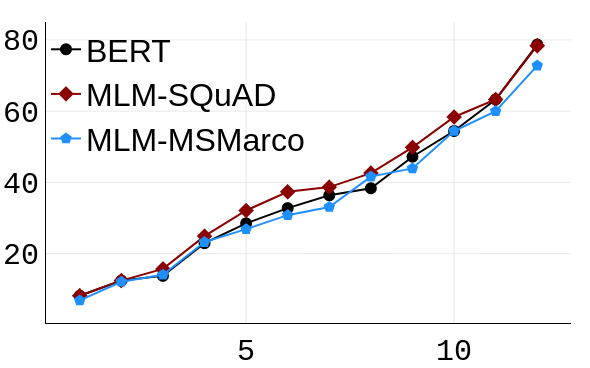}
  \caption{Squad}
\end{subfigure}
\begin{subfigure}{.24\textwidth}
  \centering
  \includegraphics[width=.99\linewidth]{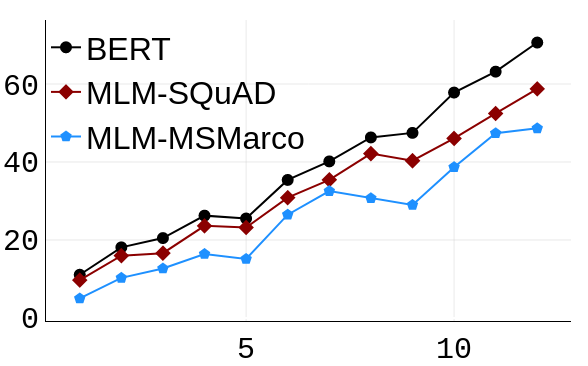}
  \caption{Google-RE}
\end{subfigure}
\caption{Effect of dataset size. Showing P@100}
\label{fig:dataset_mlm_p_100}
\end{figure*}

\begin{figure*}
\begin{subfigure}{.24\textwidth}
  \centering
  \includegraphics[width=.99\linewidth]{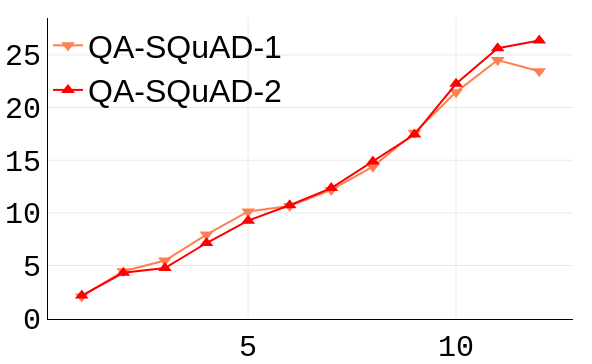}
  \caption{ConceptNet}
\end{subfigure}%
\begin{subfigure}{.24\textwidth}
  \centering
  \includegraphics[width=.99\linewidth]{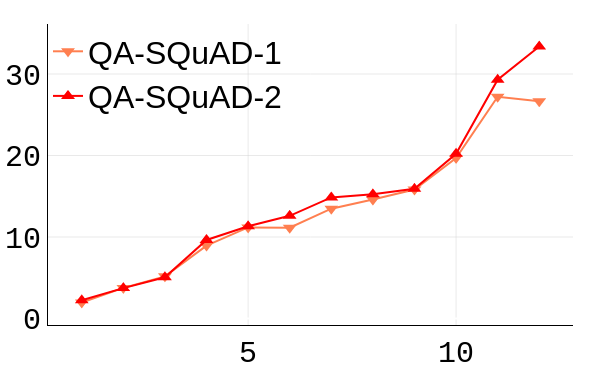}
  \caption{T-REx}
\end{subfigure}%
\begin{subfigure}{.24\textwidth}
  \centering
  \includegraphics[width=.99\linewidth]{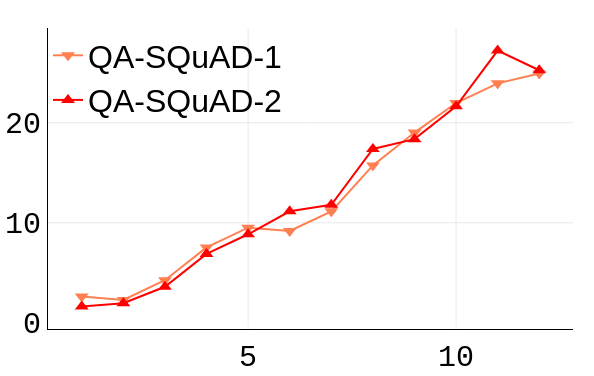}
  \caption{Squad}
\end{subfigure}
\begin{subfigure}{.24\textwidth}
  \centering
  \includegraphics[width=.99\linewidth]{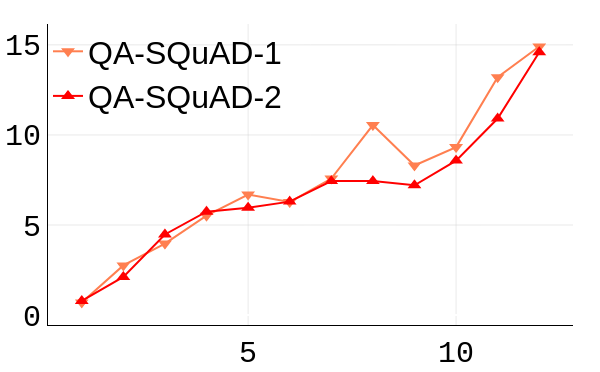}
  \caption{Google-RE}
\end{subfigure}
    \caption{Effect of dataset size. Showing P@10 for the QA objective.}
    \label{fig:dataset_qa_squad_p_10}
\end{figure*}

\begin{figure*}
\begin{subfigure}{.24\textwidth}
  \centering
  \includegraphics[width=.99\linewidth]{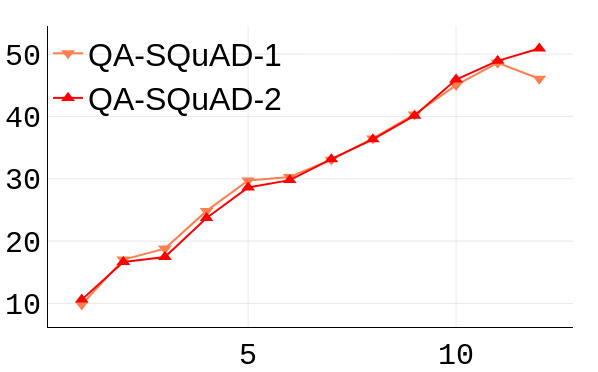}
  \caption{ConceptNet}
\end{subfigure}%
\begin{subfigure}{.24\textwidth}
  \centering
  \includegraphics[width=.99\linewidth]{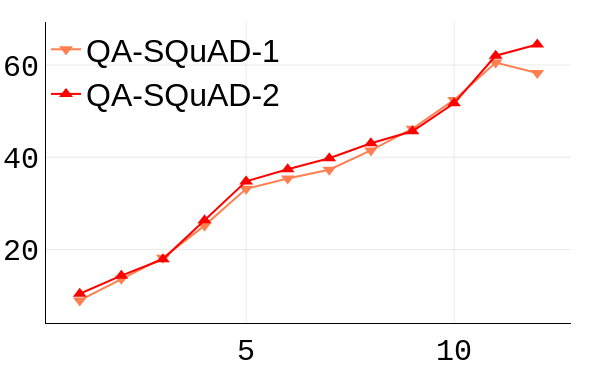}
  \caption{T-REx}
\end{subfigure}%
\begin{subfigure}{.24\textwidth}
  \centering
  \includegraphics[width=.99\linewidth]{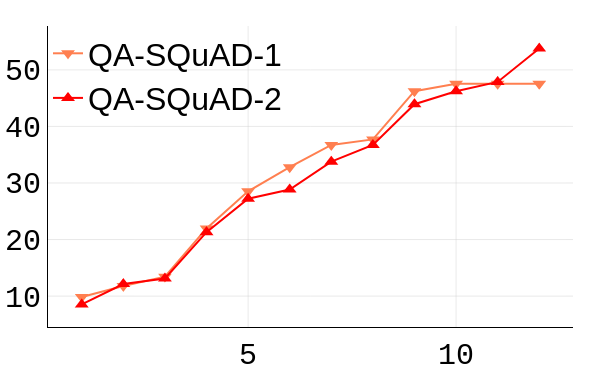}
  \caption{Squad}
\end{subfigure}
\begin{subfigure}{.24\textwidth}
  \centering
  \includegraphics[width=.99\linewidth]{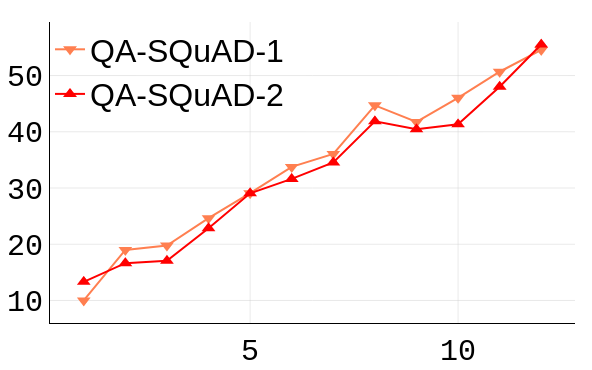}
  \caption{Google-RE}
\end{subfigure}
    \caption{Effect of dataset size. Showing P@100 for the QA objective.}
    \label{fig:dataset_qa_squad_p_100}
\end{figure*}

Figure \ref{fig:dataset_mlm_p_10} and \ref{fig:dataset_mlm_p_100} show the P@10 and P@100 plots for Figure \ref{fig:dataset_mlm}. Respectively, Figure \ref{fig:dataset_qa_squad_p_10} and \ref{fig:dataset_qa_squad_p_100} show the same for \ref{fig:dataset_qa_squad}.

\subsection{Effect of fine tuning objective}
\begin{figure*}
\begin{subfigure}{.24\textwidth}
  \centering
  \includegraphics[width=.99\linewidth]{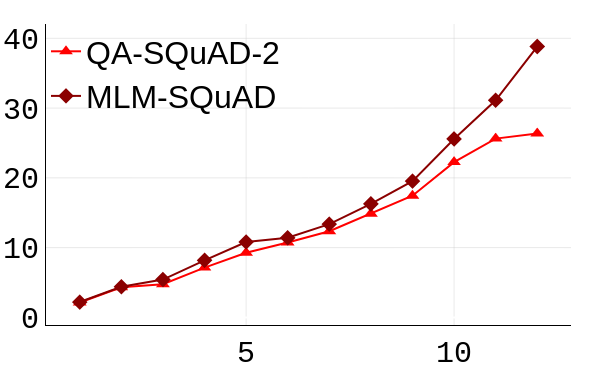}
  \caption{ConceptNet}
\end{subfigure}%
\begin{subfigure}{.24\textwidth}
  \centering
  \includegraphics[width=.99\linewidth]{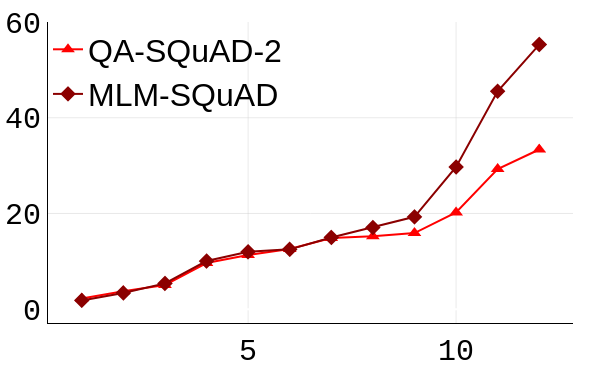}
  \caption{T-REx}
\end{subfigure}%
\begin{subfigure}{.24\textwidth}
  \centering
  \includegraphics[width=.99\linewidth]{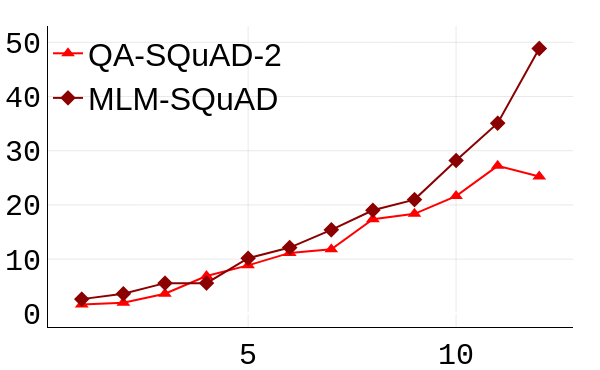}
  \caption{Squad}
\end{subfigure}
\begin{subfigure}{.24\textwidth}
  \centering
  \includegraphics[width=.99\linewidth]{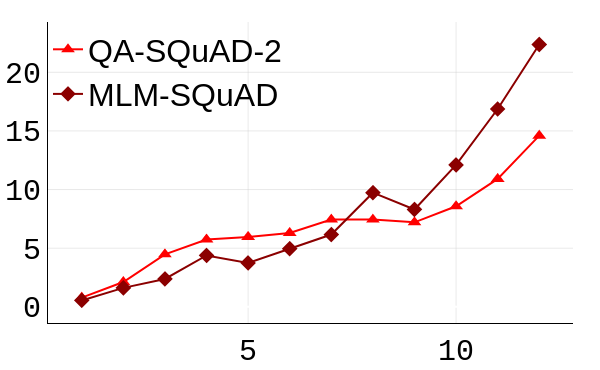}
  \caption{Google-RE}
\end{subfigure}
    \caption{Effect of Fine-Tuning Objective on fixed size data: SQUAD. Showing P@10.}
    \label{fig:finetuning_squad_p_10}
\end{figure*}

\begin{figure*}
\begin{subfigure}{.24\textwidth}
  \centering
  \includegraphics[width=.99\linewidth]{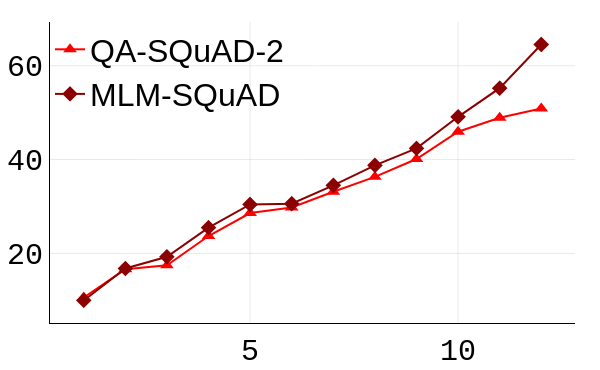}
  \caption{ConceptNet}
\end{subfigure}%
\begin{subfigure}{.24\textwidth}
  \centering
  \includegraphics[width=.99\linewidth]{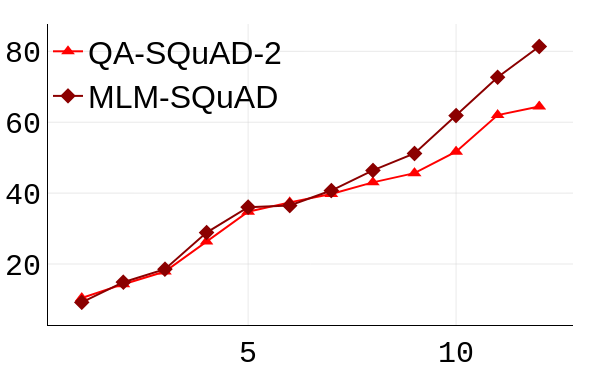}
  \caption{T-REx}
\end{subfigure}%
\begin{subfigure}{.24\textwidth}
  \centering
  \includegraphics[width=.99\linewidth]{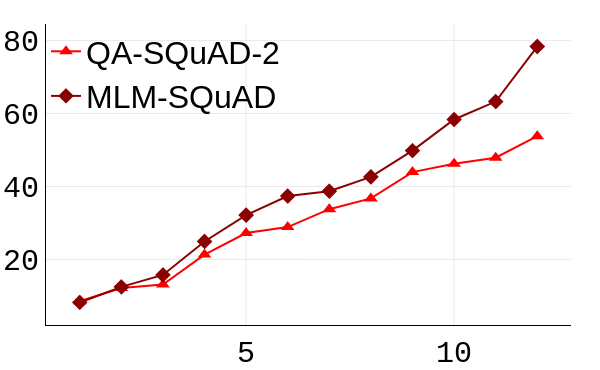}
  \caption{Squad}
\end{subfigure}
\begin{subfigure}{.24\textwidth}
  \centering
  \includegraphics[width=.99\linewidth]{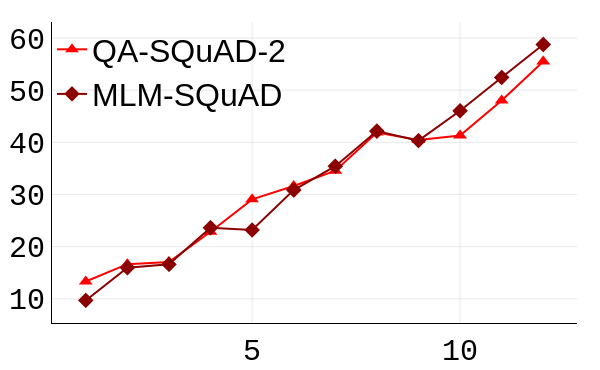}
  \caption{Google-RE}
\end{subfigure}
    \caption{Effect of Fine-Tuning Objective on fixed size data: SQUAD. Showing P@100.}
    \label{fig:finetuning_squad_p_100}
\end{figure*}

\begin{figure*}
\begin{subfigure}{.24\textwidth}
  \centering
  \includegraphics[width=.99\linewidth]{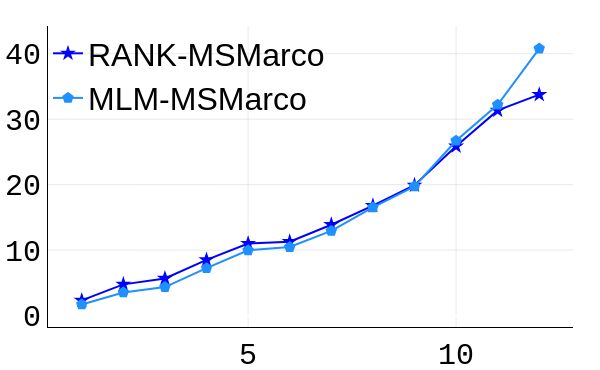}
  \caption{ConceptNet}
\end{subfigure}%
\begin{subfigure}{.24\textwidth}
  \centering
  \includegraphics[width=.99\linewidth]{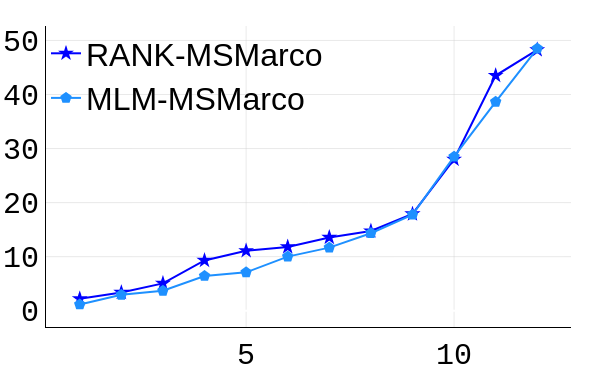}
  \caption{T-REx}
\end{subfigure}%
\begin{subfigure}{.24\textwidth}
  \centering
  \includegraphics[width=.99\linewidth]{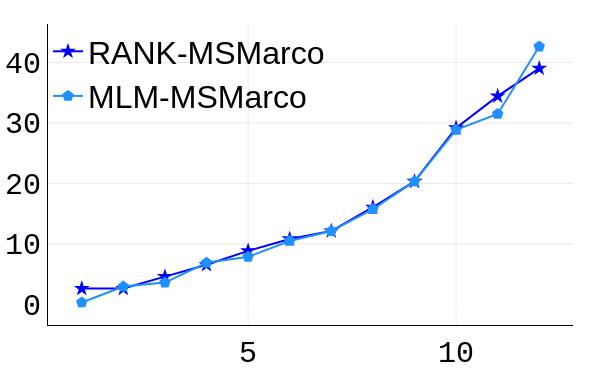}
  \caption{Squad}
\end{subfigure}
\begin{subfigure}{.24\textwidth}
  \centering
  \includegraphics[width=.99\linewidth]{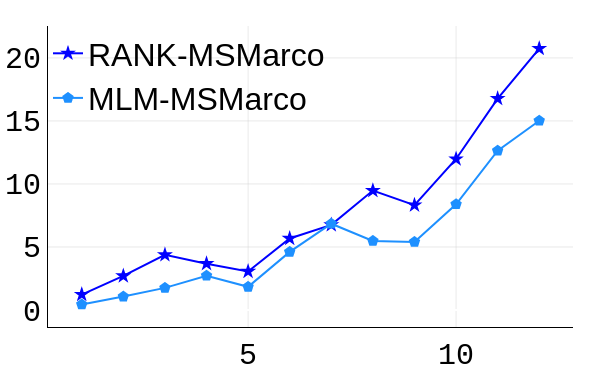}
  \caption{Google-RE}
\end{subfigure}
    \caption{Effect of Fine-Tuning Objective on fixed size data: MSMarco. Showing P@10.}
    \label{fig:finetuning_msm_p_10}
\end{figure*}

\begin{figure*}
\begin{subfigure}{.24\textwidth}
  \centering
  \includegraphics[width=.99\linewidth]{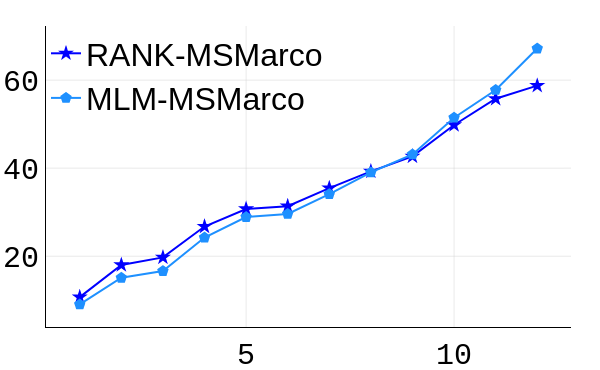}
  \caption{ConceptNet}
\end{subfigure}%
\begin{subfigure}{.24\textwidth}
  \centering
  \includegraphics[width=.99\linewidth]{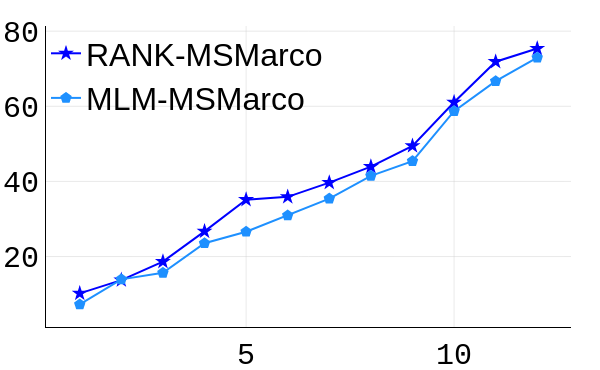}
  \caption{T-REx}
\end{subfigure}%
\begin{subfigure}{.24\textwidth}
  \centering
  \includegraphics[width=.99\linewidth]{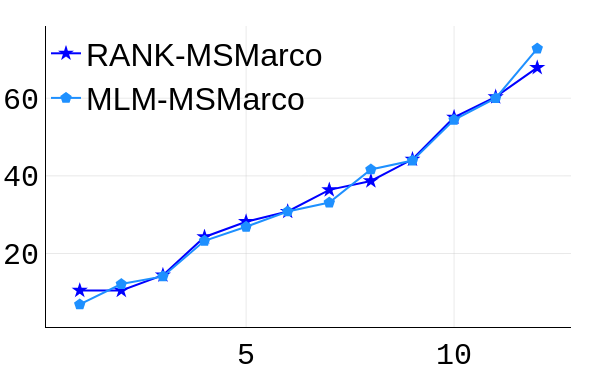}
  \caption{Squad}
\end{subfigure}
\begin{subfigure}{.24\textwidth}
  \centering
  \includegraphics[width=.99\linewidth]{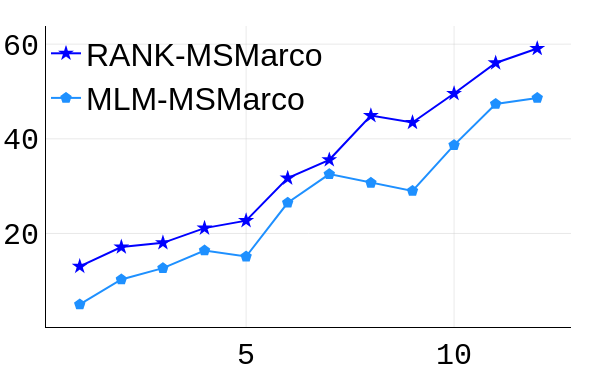}
  \caption{Google-RE}
\end{subfigure}
    \caption{Effect of Fine-Tuning Objective on fixed size data: MSMarco. Showing P@100.}
    \label{fig:finetuning_msm_p_100}
\end{figure*}

For comparing MLM and QA on SQuAD (\ref{fig:finetuning_squad}), Figure \ref{fig:finetuning_squad_p_10} and \ref{fig:finetuning_squad_p_100} show more precisions. Also, for comparing fine tune objectives on MSMARCO (Figure \ref{fig:finetuning_msm}), Figure \ref{fig:finetuning_msm_p_10} and \ref{fig:finetuning_msm_p_100} show P@10 and P@100. 

\subsection{MLM and learning factual knowledge}

\begin{table}[]
\begin{tabular}{lccccc}
\hline
\textbf{Model}      & \multicolumn{1}{l}{\textbf{Google-RE}} & \multicolumn{1}{l}{\textbf{T-REx}} & \multicolumn{1}{l}{\textbf{ConceptNet}} & \multicolumn{1}{l}{\textbf{Squad}} & \multicolumn{1}{l}{\textbf{MLM loss}} \\ \hline
BERT                & 0.10                                   & 0.29                               & 0.15                                    & 0.13                               & 2.115                                 \\ \hline
Evidences-object-1  & 0.10                                   & 0.34                               & 0.39                                    & 0.43                               & 3.156 (+49\%)                         \\
Evidences-object-2  & 0.10                                   & 0.30                               & 0.54                                    & 0.86                               & 4.18 (+98\%)                          \\
Evidences-object-10 & 0.17                                   & 0.36                               & 0.86                                    & 0.99                               & 6.041 (+184\%)                        \\ \hline
Evidences-random-1  & 0.06                                   & 0.26                               & 0.20                                    & 0.16                               & 2.843 (+34\%)                         \\
Evidences-random-2  & 0.09                                   & 0.26                               & 0.19                                    & 0.20                               & 3.634 (+72\%)                         \\
Evidences-random-10 & 0.08                                   & 0.26                               & 0.32                                    & 0.25                               & 4.863 (+130\%)                        \\ \hline
\end{tabular}
\caption{Capacity results from training 1,2,10 epochs on the probing data. The training dataset is not shuffled and it is trained sequentially in the same order as the table: \greprobe{}, \trexprobe{}, \concpetprobe{}, \squadprobe{}. We used the evidence sentences (see Table~\ref{tab:facts_template_evidence}) and either masked objects or random tokens.}
\label{tab:mlm_evidence_random_vs_objects}
\end{table}

\end{document}